\documentclass[10pt,twocolumn,letterpaper]{article}

\usepackage{wacv}
\usepackage{times}
\usepackage{epsfig}
\usepackage{graphicx}
\usepackage{amsmath}
\usepackage{amssymb}

\usepackage{paralist}
\usepackage{booktabs}
\usepackage{subcaption}
\usepackage[sort,nocompress]{cite}
\usepackage[accsupp]{axessibility}  

\def\wacvPaperID{128}

\wacvapplicationstrack

\wacvfinalcopy

\ifwacvfinal
\usepackage[breaklinks=true,colorlinks,bookmarks=false]{hyperref}
\else
\usepackage[pagebackref=true,breaklinks=true,colorlinks,bookmarks=false]{hyperref}
\fi

\DeclareGraphicsExtensions{.pdf,.jpg,.png}
\graphicspath{{figs/}}

\newcommand{\secref}[1]{Section~\ref{sec:#1}}
\newcommand{\figref}[1]{Figure~\ref{fig:#1}}
\newcommand{\tabref}[1]{Table~\ref{tbl:#1}}

\makeatletter
\renewcommand{\paragraph}{%
  \@startsection{paragraph}{4}%
  {\z@}{1.5ex \@plus 1ex \@minus .2ex}{-1em}%
  {\normalfont\normalsize\bfseries}%
}
\makeatother

\begin{document}

\title{Handling Image and Label Resolution Mismatch in Remote Sensing}

\author{
  \begin{minipage}{.9\linewidth}
    \centering
    \begin{minipage}{1.6in}
      \centering
      Scott Workman
      \\[.15cm]
      \normalsize{DZYNE Technologies}
    \end{minipage}
    \begin{minipage}{1.6in}
      \centering
      Armin Hadzic
      \\[.15cm]
      \normalsize{DZYNE Technologies}
    \end{minipage}
    \begin{minipage}{1.6in}
      \centering
      M. Usman Rafique
      \\[.15cm]
      \normalsize{Kitware Inc.}
    \end{minipage}
  \end{minipage}
}

\maketitle

\begin{abstract}

    Though semantic segmentation has been heavily explored in vision literature, unique challenges remain in the remote sensing domain. One such challenge is how to handle resolution mismatch between overhead imagery and ground-truth label sources, due to differences in ground sample distance. To illustrate this problem, we introduce a new dataset and use it to showcase weaknesses inherent in existing strategies that naively upsample the target label to match the image resolution. Instead, we present a method that is supervised using low-resolution labels (without upsampling), but takes advantage of an exemplar set of high-resolution labels to guide the learning process. Our method incorporates region aggregation, adversarial learning, and self-supervised pretraining to generate fine-grained predictions, without requiring high-resolution annotations. Extensive experiments demonstrate the real-world applicability of our approach.

\end{abstract}

\section{Introduction}

Semantic segmentation is a fundamental computer vision problem where the goal is to assign each individual pixel of an image to a semantic class. This research area has been heavily explored for decades and is critical for many applications, such as autonomous driving~\cite{siam2018comparative,feng2020deep}. Recently, advances in machine learning have pushed performance levels to new heights. However, despite the success of these methods when applied to ground-level imagery, there remain numerous challenges to successfully applying semantic segmentation to imagery from the remote sensing domain~\cite{hua2021semantic}.

\begin{figure}
    \centering
    \begin{subfigure}[b]{.49\linewidth}
        \includegraphics[width=1\linewidth]{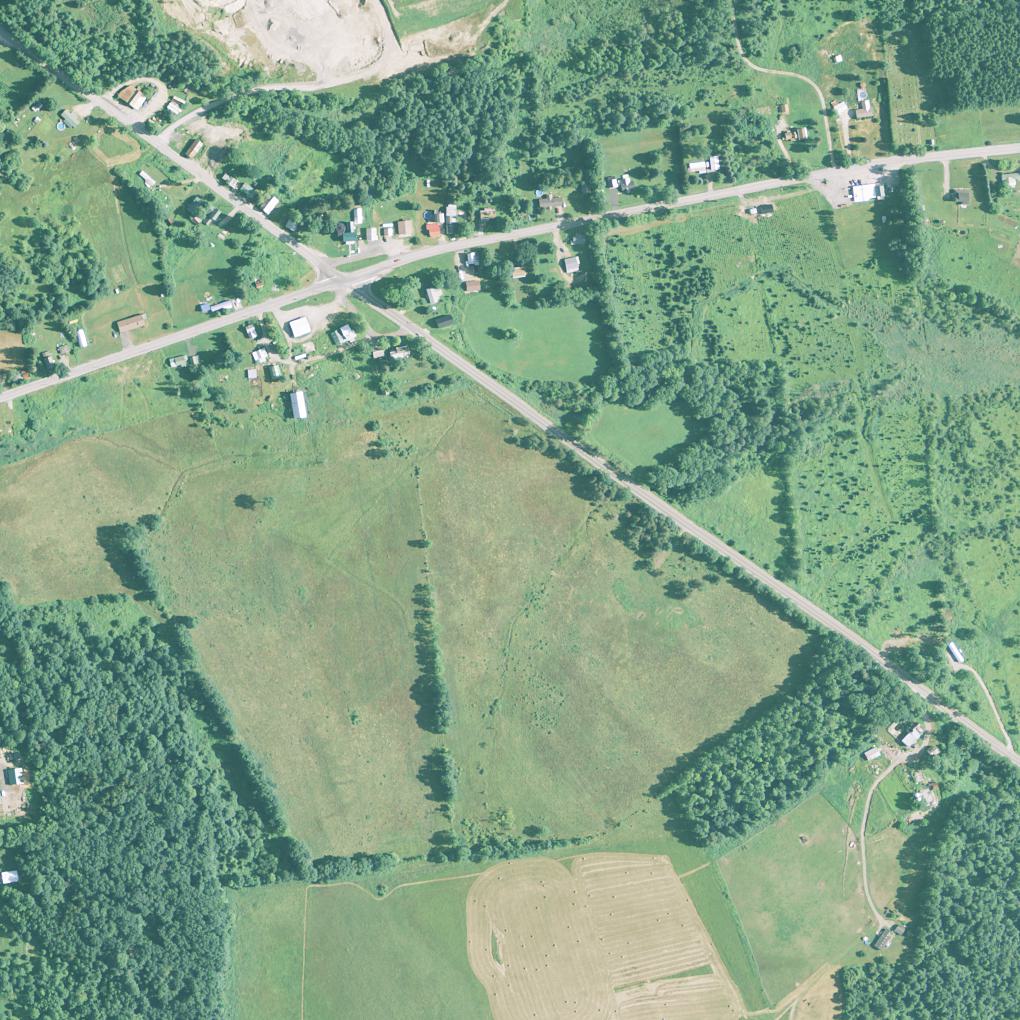}
        \caption{Image}
    \end{subfigure}
    \begin{subfigure}[b]{.49\linewidth}
        \includegraphics[width=1\linewidth]{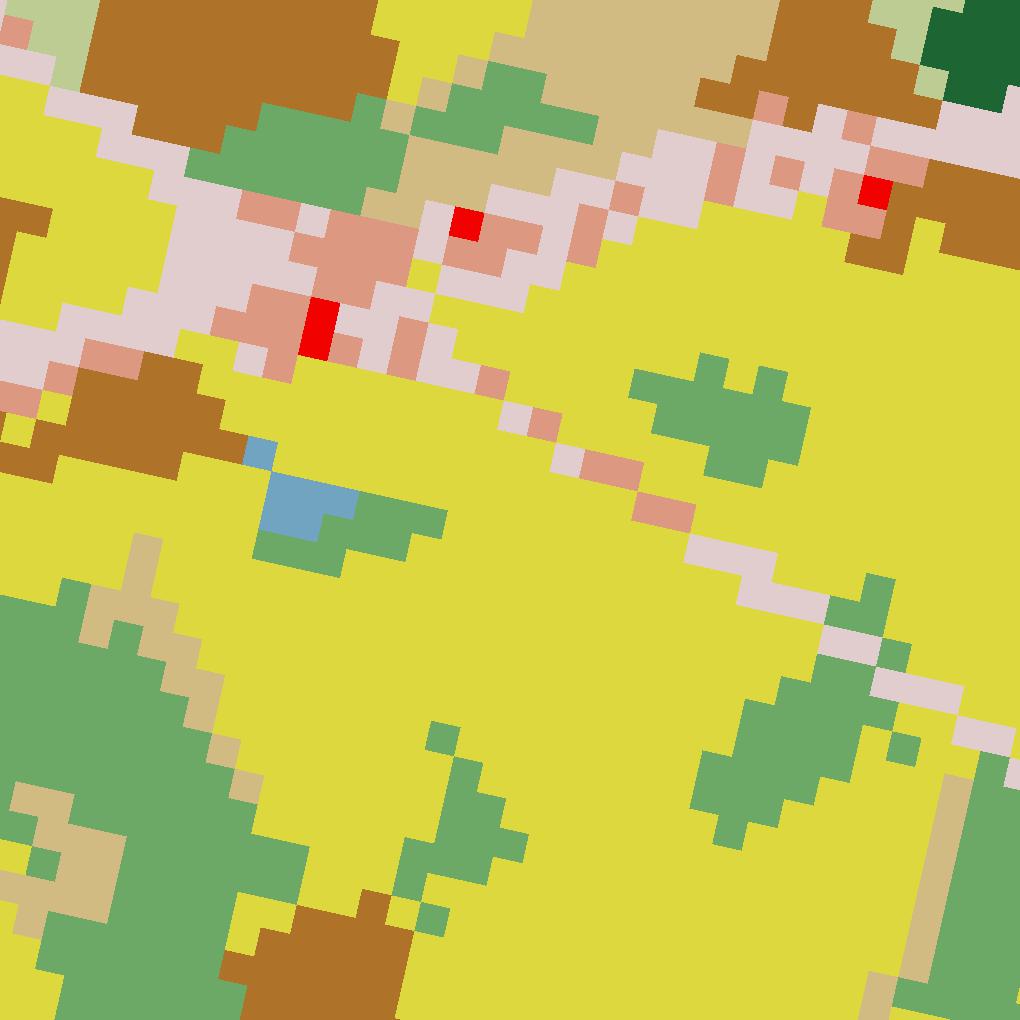}
        \caption{Label}
    \end{subfigure}
    
    \caption{Semantic segmentation in remote sensing has many unique challenges, such as differences in spatial resolution between overhead imagery and target labels. This is primarily due to the increased resource cost of collecting fine-grained annotations at high-resolutions. We propose a method for handling this resolution mismatch, without the need for high-resolution annotations.}
    \label{fig:cartoon}
\end{figure}

Central to the issue is that overhead imagery comes in many diverse formats. Considering space-based remote sensing alone, there are hundreds of different imaging sensors orbiting the earth, each capturing information in different ways. These sensors can have different imaging modalities (e.g., multispectral, radar), measure light differently, varying ground resolutions (also known as ground sample distance), unique look angles in relation to the target, and much more. Beyond capture details, the appearance of a scene can vary drastically for many reasons, including seasonal variations and artifacts such as clouds and cloud shadows. Many of these issues are unique to remote sensing and require novel solutions.

The diversity of overhead imagery creates downstream problems for semantic segmentation algorithms. For example, given off-nadir imagery (i.e., not captured from directly overhead), there is often misalignment between the imagery and ground-truth labels. There is a large body of work exploring how to address related issues in remote sensing. For example, Christie et al.~\cite{christie2021single} propose a method to regress the geocentric pose of an overhead image and show its utility for building segmentation and rectification. Deng et al.~\cite{deng2021scale} propose a framework for handling differences in scale between overhead image datasets. In this work, we focus our efforts on how to handle resolution mismatch between overhead imagery and target annotations, a relatively unexplored problem (\figref{cartoon}).

Semantic segmentation depends heavily on having high quality, aligned labels. A recent study by Zlateski et al.~\cite{zlateski2018importance} found that segmentation performance primarily depends on the time spent creating training labels. With overhead imagery, high-resolution labels do not exist at a large scale. This is primarily due to practical limitations associated with the enormous resource costs needed for comprehensive annotation efforts. For example, the Chesapeake Conservancy spent 10 months and \$1.3 million to produce a high-resolution (1 meter) land cover map of the Chesapeake Bay watershed~\cite{robinson2019large}. The result is a land cover dataset for a small portion of the globe, with a fixed spatial resolution, at only a single timestep.

In practice, it is often significantly easier to acquire low-resolution labels. For example, the National Land Cover Database~\cite{yang2018new} (NLCD) is freely available to the public, has complete coverage of the continental United States, is typically updated every five years, but has a spatial resolution of only 30 meters per pixel. In the case that an input overhead image is of higher resolution than the target label, the typical strategy is to simply upsample the label to match the native resolution of the image. Our experiments demonstrate that this approach is unsatisfactory and results in low quality output.

Instead, our goal is to develop a method capable of generating fine-grained predictions, but only using low-resolution ground-truth as a direct source of supervision. Our key insight is that even if high-resolution ground-truth is unavailable for the training imagery, examples of high-resolution annotations are often readily available for other locations. In other words, it is possible to observe what high-resolution output looks like, just not everywhere. A method should be able to take advantage of this auxiliary data, without the corresponding imagery, to aid in producing a fine-grained output.

We present a method that is supervised only using low-resolution labels, but takes advantage of an exemplar set of high-resolution labels to guide the learning process. Our method has several key components: 1) it incorporates the concept of region aggregation to allow the network to output native resolution predictions, without upsampling the low-resolution target label; 2) it uses adversarial learning combined with an exemplar set of high-resolution labels to encourage predictions to be fine-grained; and 3) it leverages self-supervised pretraining on a large set of unlabeled imagery to increase model generalization. The result is a method which bridges the performance gap between naively upsampling available low-resolution labels and assuming the existence of expensive high-resolution labels. Extensive experiments demonstrate the capability of our approach.

\begin{figure*}
    \centering
    \begin{subfigure}[b]{.195\linewidth}
        \includegraphics[width=1\linewidth]{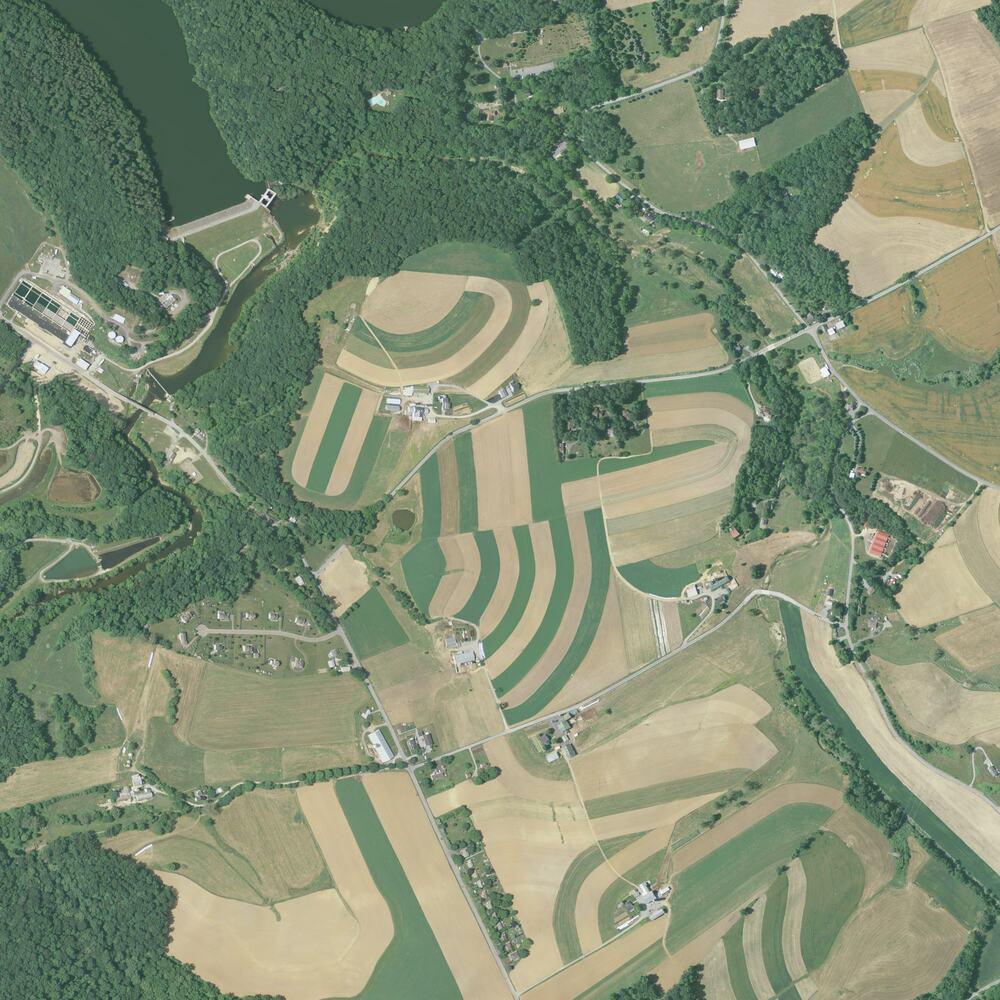}
        \caption{Image}
    \end{subfigure}
    \begin{subfigure}[b]{.195\linewidth}
        \includegraphics[width=1\linewidth]{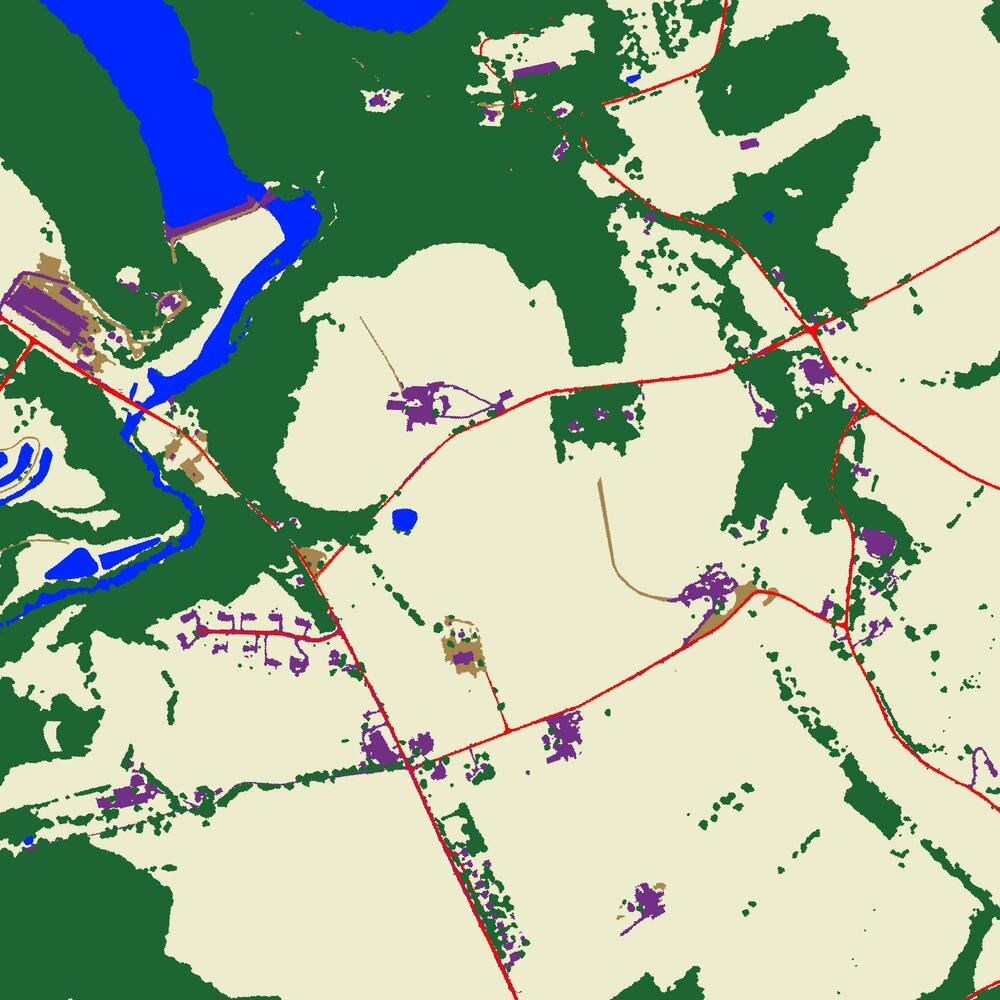}
        \caption{Chesapeake~\cite{robinson2019large}}
    \end{subfigure}
    \begin{subfigure}[b]{.195\linewidth}
        \includegraphics[width=1\linewidth]{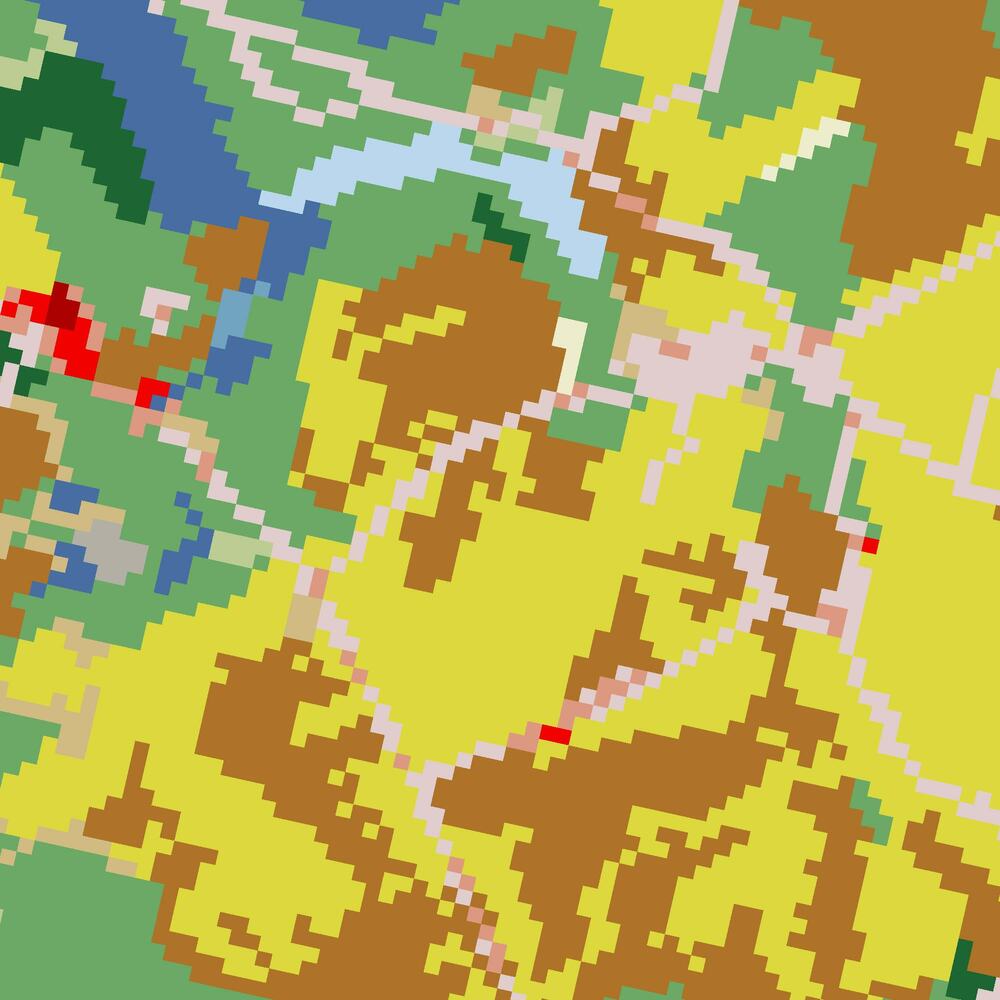}
        \caption{NLCD~\cite{yang2018new}}
    \end{subfigure}
    \begin{subfigure}[b]{.195\linewidth}
        \includegraphics[width=1\linewidth]{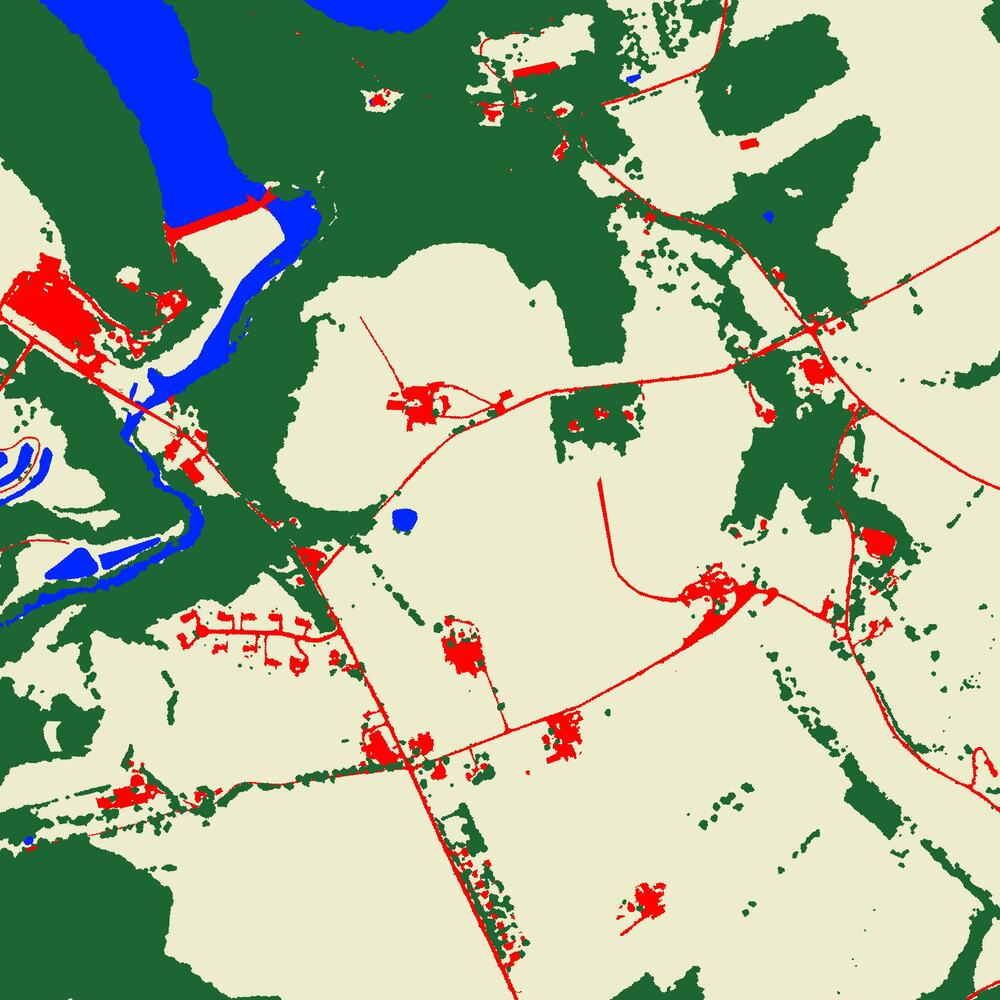}
        \caption{Chesapeake (merged)}
    \end{subfigure}
    \begin{subfigure}[b]{.195\linewidth}
        \includegraphics[width=1\linewidth]{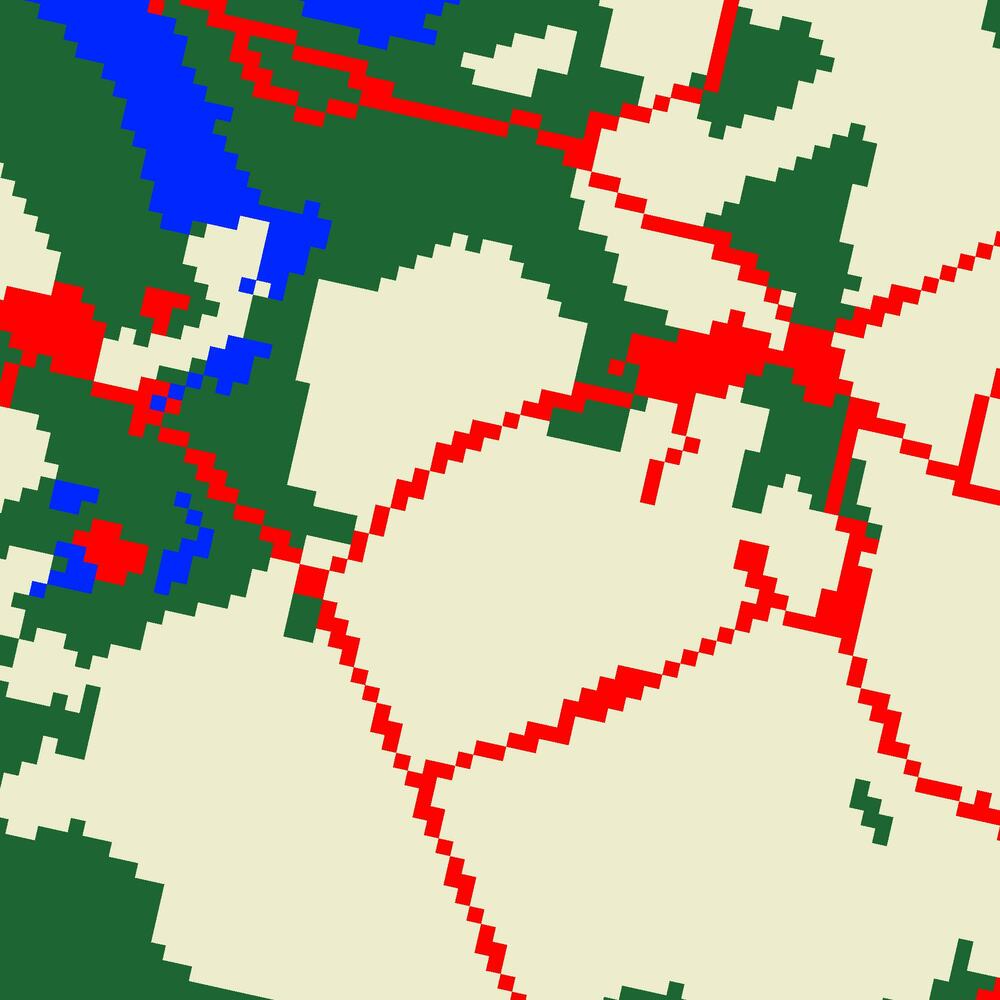}
        \caption{NLCD (merged)}
    \end{subfigure}
    \caption{Example labels using our merged label taxonomy. Note that the low-resolution label (e) omits many fine-grained details, such as structures and secondary roads, that the high-resolution label (d) captures.}
    \label{fig:quality}
\end{figure*}

\section{Related Work}

Semantic segmentation is considered a foundational task in computer vision, a necessary stepping stone towards the larger goal of intelligent scene understanding~\cite{li2022deep}. Much of the recent progress in semantic segmentation can be attributed to: 1) the development of learning-based segmentation algorithms (see \cite{minaee2021image} for a comprehensive survey) and 2) the introduction of large-scale benchmark datasets such as Cityscapes~\cite{cordts2016cityscapes}, ADE20k~\cite{zhou2017scene}, and Mapillary Vistas~\cite{neuhold2017mapillary}, which contain pixel-wise annotations that are important for enabling fully-supervised segmentation approaches.

Semantic segmentation has attracted a great deal of attention in the remote sensing community as well~\cite{zhu2017deep}. It is apt for traditional remote sensing tasks such as land cover and land use estimation~\cite{vali2020deep}, which seek to understand the physical cover of the Earth's surface and how it is utilized. M{\'a}ttyus et al.~\cite{mattyus2017deeproadmapper} segment roads in overhead imagery as part of a method for directly extracting road topology. Girard et al.~\cite{girard2021polygonal} leverage semantic segmentation to extract building polygons. Other applications include height estimation~\cite{mahmud2020boundary,workman2021augmenting}, road safety assessment~\cite{song2018farsa}, traffic monitoring~\cite{workman2020dynamic, hadzic2020rasternet}, emissions estimation~\cite{mukherjee2021towards}, near/remote sensing~\cite{workman2017unified,workman2022revisiting}, and more. However, similar to ground-level imagery, a major difficulty in applying semantic segmentation to overhead imagery is the cost of acquiring annotated training data~\cite{hua2021semantic}.

Remote-sensed imagery presents many unique obstacles to contend with. Imagery can come from many different sensors, have different spatial resolutions, contain atmospheric artifacts such as clouds, etc. Deng et al.~\cite{deng2021scale} propose a method for handling the scale variation commonly found between remote sensing image collections and demonstrate its application to domain adaptation. Workman et al.~\cite{workman2020single} show how multi-image fusion can be applied to detecting artifacts in imagery, such as clouds. These issues, and more, compound the difficulty of getting annotated training data. Ultimately, it is common for annotations to have a different spatial resolution, be captured at a different time, have spatial alignment errors, or a slew of other discrepancies.

In this work, we focus specifically on the resolution mismatch problem. As high-resolution imagery has become more widely available, for example WorldView-3 imagery at approximately 30 centimeters, it has become increasingly difficult to acquire high-resolution annotations to match the native image resolution. Instead, the typical strategy is to simply upsample the target label to match the image resolution. The 2020 IEEE GRSS Data Fusion Contest (DFC2020)~\cite{yokoya20202020} was of the first to consider the problem of resolution mismatch, but at an extreme scale (high-resolution labels of 10 meters and low-resolution labels of 500 meters), necessitating non-learning based methods. Robinson et al.~\cite{robinson2020weakly} use an ensemble of individual solutions including: label super-resolution via iterative clustering, label super-resolution using  epitomic representations~\cite{malkin2020mining}, and post-processing inspired by deep image prior~\cite{ulyanov2018deep}. Instead, we explore an end-to-end learning based approach for addressing the mismatch problem by generating a high-resolution prediction in a single forward pass without an ensemble or any ad hoc post-processing steps.

Several weakly supervised methods have been proposed to take advantage of sparse ground-truth labels. Lu et al.~\cite{lu2021point} show how geotagged point labels can be used to identify bodies of water via the introduction of a feature aggregation network. Wang et al.~\cite{wang2020weakly} also consider geotagged points, as well as image-level labels, and show that standard segmentation architectures can outperform pixel-level algorithms. We consider an alternative, more common scenario, where annotations are dense but of a lower spatial resolution than the input imagery.

\section{Low2High: A Dataset for Resolution Mismatch}

To support our experiments, we introduce the Low2High dataset which extends the recent Chesapeake dataset~\cite{robinson2019large}. The Chesapeake dataset contains high-resolution overhead imagery, at 1m per pixel, from the USDA National Agriculture Imagery Program (NAIP). In total, the dataset consists of over 700 non-overlapping tiles from six states, with each tile corresponding to an area of approximately 6km $\times$ 7.5km. This imagery is combined with land cover labels from two different sources. The first source is high-resolution labels (1m per pixel) obtained from the Chesapeake Conservancy land cover project covering approximately 100,000 square miles of land around the Chesapeake Bay watershed. The second source is low-resolution labels (30m per pixel) from the National Land Cover Database (NLCD). We extend this dataset to support our experiments related to the resolution mismatch problem.

\begin{figure*}
    \centering
    \includegraphics[width=.9\linewidth]{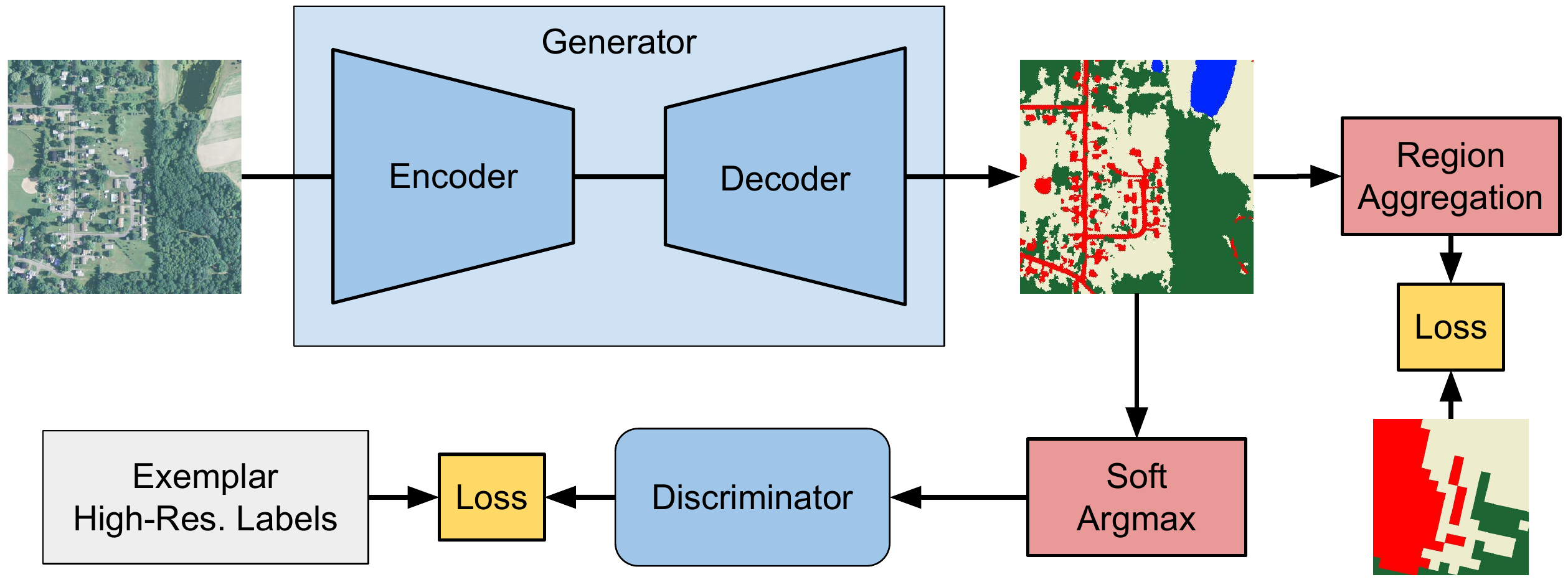}
    \caption{An overview of our architecture for handling resolution mismatch.}
    \label{fig:architecture}
\end{figure*}

\subsection{Dataset Generation}

We start from the raw tiles in the Chesapeake dataset, update the land cover labels to use a merged label taxonomy, and generate an aligned dataset of non-overlapping images of size $512 \times 512$. We promote label diversity and omit any image with less than 3 unique land cover classes (either label source), or that has a single land cover class covering more than 75 percent of the image. This process results in $34791$ images. We also generate a custom evaluation split, including a held-out set of images from each state. The final split consists of training (75\%), validation (10\%), held-out (10\%), and testing (5\%).

\paragraph{Merged Label Taxonomy}
The high-resolution land cover labels from the Chesapeake Conservancy include six land cover classes: tree canopy / forest, low vegetation / field, barren land, impervious (other), and impervious (road). The low-resolution land cover labels from NLCD include twenty land cover classes with high-level categories spanning developed, barren, forest, and shrub land. For our purposes, we generate a merged label taxonomy consisting of four classes: water, forest, field, and impervious. Each label source is remapped to use the merged label taxonomy by assigning existing labels to the most relevant category (see the supplemental material for details). \figref{quality} shows qualitative examples of this process. Though the labels often tend to agree there can be stark differences, with the low-resolution label missing certain details completely.

\subsection{Auxiliary Imagery \& Labels}
\label{sec:auxiliary}

Self-supervised learning, where a supervisory signal is obtained directly from the data itself, is an alternative training paradigm capable of taking advantage of large amounts of unlabeled data. To support this scenario and improve geographic diversity, we supplement Low2High with an auxiliary set of images collected near rest stops across the contiguous United States~\cite{intRA}. Similar to the format of the Chesapeake imagery, we obtained overhead imagery ($512 \times 512$) from NAIP between 01/01/2013 and 12/31/2014, resulting in 12,024 training and 603 validation images.

We also supplement the Chesapeake data with a new held-out test region to support domain adaptation experiments. Milwaukee, Wisconsin was selected as the new test site due to its diverse geography compared to the Northeastern USA. For land cover labels, we use annotations made available through the EPA EnviroAtlas~\cite{pilant2020us} and, as before, the land cover classes were remapped to use our merged label taxonomy. The resulting test set contains 3262 samples at 1m ground sample distance.

\section{A Method for Semantic Segmentation with Resolution Mismatch}
\label{sec:method}

We propose an end-to-end architecture for semantic segmentation that is supervised using low-resolution annotations, but is capable of generating fine-grained predictions. \figref{architecture} provides a visual overview of our approach.

\subsection{Approach Overview}

We present a framework for semantic segmentation in the event of resolution mismatch. Our method is directly supervised with low-resolution labels, but takes advantage of an exemplar set of high-resolution labels (no correspondence to the input imagery) to guide the learning process. Our architecture has four primary components. First, we describe the general segmentation architecture (\secref{segmentation}). Second, we incorporate the concept of region aggregation to allow the network to output native resolution predictions, without upsampling the low-resolution target label (\secref{aggregate}). Third, we use adversarial learning combined with an exemplar set of high-resolution labels to encourage predictions to be fine-grained (\secref{adversarial}). Finally, we leverage self-supervised pretraining on a large set of unlabeled imagery to increase model generalization (\secref{self-supervised}). The remainder of this section describes the high-level architecture. See the supplemental materials for additional details.

\subsection{Segmentation Architecture}
\label{sec:segmentation}

For our segmentation architecture, we use a variant of U-Net~\cite{ronneberger2015u} with a ResNet-18 backbone. However, our approach is general enough to be conceivably combined with any modern segmentation architecture. U-Net is an encoder-decoder style architecture that propagates information from the encoder through the use of skip connections. For our feature encodings, we use the output activations from the first four stages of ResNet. The decoder (U-Net style) expects four feature maps to be used for skip connections. Our variant has four upsampling blocks, each consisting of two convolutions ($3\times3$) followed by a ReLU activation. The output is then upsampled and passed through a final convolutional layer with number of output channels that are task-specific. In our case, the segmentation architecture takes an overhead image as input and produces a native resolution output.

\subsection{Region Aggregation}
\label{sec:aggregate}

The typical strategy in the event of resolution mismatch between input image and target label is to simply upsample the label to the resolution of the input. Our experiments demonstrate that this approach is sub-optimal. Instead, we use a variant of region aggregation~\cite{jacobs2018weakly, workman2020dynamic} to allow our network to generate native resolution outputs, yet be supervised with low-resolution target labels (without any upsampling). This allows the underlying segmentation network to generate fine-grained predictions.

\begin{figure}
    \includegraphics[trim={0 .4cm .4cm .3cm},clip,width=1\linewidth]{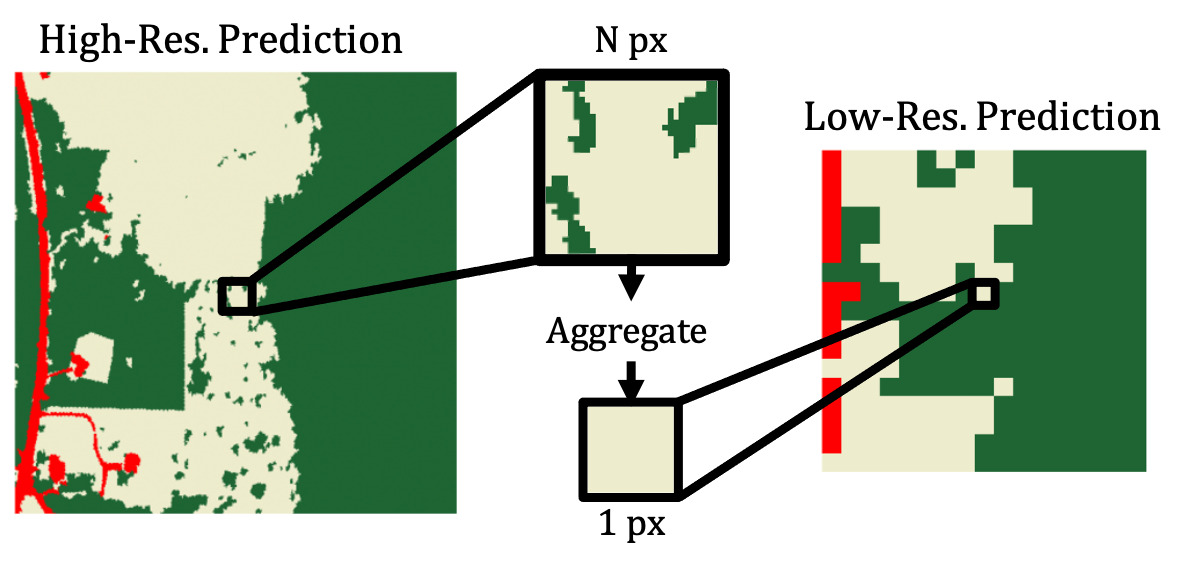}
    \caption{Our region aggregation component takes advantage of the known geospatial context of an input image to aggregate high-resolution predictions such that they match the spatial resolution of a low-resolution target label.}
    \label{fig:region_agg}
\end{figure}

In the case of semantic segmentation of overhead imagery, the input imagery and target label are both georeferenced. This means that the ground sample distance of each is known, and enables computation of the geospatial correspondence between a single low-resolution pixel (from the target label) and many high-resolution prediction pixels (from the input image). See \figref{region_agg} for a visual illustration of this geospatial relationship. The ratio of high-resolution pixels to low-resolution pixels ($n$) can be directly computed from the respective ground sample distances.

We construct a region index matrix $M\in \mathbb{R}^{H \times W}$, which expresses a pixel mapping from high-resolution to low-resolution across a given image of height $H$ and width $W$. $M$ is composed of $t$ number of $n \times n$ pixel regions, where $n$ is the ground sample distance scale-ratio from the high-resolution imagery to the low-resolution labels. For the case of mapping 1m NAIP imagery to 30m NLCD annotations, $n=30$. We use $M$ to select $30 \times 30$ pixel regions from the high-resolution prediction, and sum them to produce a single value representative of the region. In practice, this operation is applied directly to the logits. This region-based summation operation results in a low-resolution prediction map. We then forward the low-resolution prediction map to a cross entropy loss:
\begin{equation}
    L(Y, \hat{Y}) = - \frac{1}{N}\sum_{i=1}^{N}log \left (\frac{e^{y_i}}{\sum^C_{c=1} e^{\hat{y}_{i,c}}} \right).
\end{equation}
This allows newly down-scaled low-resolution predictions to be supervised by low-resolution annotations (bottom right of \figref{architecture}). For the multi-class ($c \in C$) cross entropy loss, $y_i \in Y$ indicates that $N$ low-resolution labels were used to supervise $\hat{y} \in \hat{Y}$ aggregated predictions.

\subsection{Adversarial Learning}
\label{sec:adversarial}

We use adversarial learning combined with an exemplar set of high-resolution labels to encourage predictions to be fine-grained. In other words, we treat the segmentation network (\secref{segmentation}) as a generator ($G$). The predictions from the generator are encouraged to match the appearance of an unrelated high-resolution annotation by having the discriminator ($D$) predict whether an input is fake (a prediction) or real (a randomly selected high-resolution annotation). This is represented as a two-player minmax game:
\begin{equation}
    \min\limits_G \max\limits_D L_{adv} (D, G).
\end{equation}

To facilitate this training paradigm, predictions from the generator need to match the characteristics of randomly selected high-resolution annotations taken from a small held-out dataset. In our case, the discriminator operates on segmentation maps (indexed images) as input. The generator outputs logits ($B \times C \times H \times W$), yet the discriminator expects inputs to be in the form of class indices ($B\times 1 \times H \times W$). Transforming logits to class indices is usually addressed by applying a $softmax$ across the channels dimension followed by an $argmax$ operation. However, $argmax$ is non-differentiable, which would interrupt gradient flow from the discriminator to the generator during adversarial training.

Instead, we present a novel formulation using a differentiable soft $argmax$ operation, denoted as $sargmax$. First, class probabilities are extracted from the input logits $x$ by multiplying the logits by a temperature scaling parameter, $\tau=10^3$, and applying a $softmax$ operation across the channels dimension ($c \in C$) of $x$, indicated as $x_c$. The $sargmax$ can be expressed as:
\begin{equation}
    sargmax := \sum^C_{c=1} c * p(x_c; \tau),
\end{equation}
where class probabilities, $p(x_c;\tau)=softmax(\tau x_c)$, are subsequently passed to an expectation to produce an approximate class index per spatial position, corresponding to the highest predicted class probability. This formulation allows soft class assignments to each pixel of a segmentation output. Predictions and high-resolution exemplars are then passed to the discriminator, $D$, which will try to determine whether a given input is a real high-resolution annotation or a prediction synthesized by $G$. We follow this with the hinge loss~\cite{lim2017geometric}, $L_{adv}$, to jointly optimize the generator and discriminator.

Kendall et al.~\cite{kendall2017end} proposed a similar operation for differentiable aggregation denoted as $soft argmin$. The key distinction between our $sargmax$ and $soft argmin$ is that we use a temperature parameter $\tau$ to make the predicted probability distributions have more pronounced peaks. Distinct peaks are desirable for allowing predictions to emulate class indices, as opposed to the more common trend of using $soft argmin$ for regression.

\subsection{Self-supervised Pretraining}
\label{sec:self-supervised}

Self-supervised learning has shown great promise for a variety of tasks~\cite{chen2020simple, li2020prototypical}, including land cover segmentation~\cite{scheibenreif2022self}. The premise is to leverage various pretext tasks, which create supervisory signals from the data itself, for learning useful feature representations which can then be transferred to the downstream task. Masked Autoencoders (MAE)~\cite{he2022masked} introduced image reconstruction as a pretext task, showing that transfer performance on several downstream tasks  outperformed supervised pretraining. Inspired by these promising results, we use the MAE reconstruction task to pretrain a Vision Transformer ViT~\cite{dosovitskiy2020image} with a ResNet-18 patch embedding (masking ratio of 75\%) followed by $2\times$ upsampling layer. The weights from the pretrained ResNet-18 embedding network are then used to initialize the feature backbone of our segmentation network (\secref{segmentation}).

\subsection{Implementation Details}

We implement our methods using Pytorch~\cite{paszke2019pytorch} and Pytorch Lightning~\cite{falcon2019pytorch}. Our networks are optimized using Adam~\cite{kingma2014adam} with the initial learning rate set to $1e^{-4}$. The method is trained for 100 epochs and a validation set is used for model selection. For the adversarial learning component, the inputs passed to the discriminator are blurred using a Gaussian filter ($3 \times 3$ kernel with $\sigma=0.6$).

\section{Experiments}

We evaluate our methods quantitatively and qualitatively through a variety of experiments. Results demonstrate that our approach which incorporates region aggregation, adversarial learning, and self-supervised pretraining, significantly reduces error compared to baselines.

\paragraph{Baseline Methods}

For evaluating our proposed architecture, we consider several baseline methods that share components of our full approach:
\begin{compactitem}
    \item {\em Oracle} uses the upsampled low-resolution ground-truth directly as the prediction. This represents having a perfect low-resolution estimator.
    \item {\em Low} is trained on the low-resolution ground-truth and naively upsamples to the native image resolution.
    \item {\em High} is trained on the high-resolution ground truth. This represents ideal performance in the event high-resolution ground-truth is always available.
\end{compactitem}
Our full approach is outlined in \secref{method} and is subsequently referred to as {\em Ours}. We also compare against the label super-resolution approach, {\em Self-Epitomic LSR}~\cite{malkin2020mining}.

\paragraph{Metrics}

In our experiments, we evaluate all methods using the high-resolution labels. In other words, the goal is to examine how well each method performs relative to the high-resolution baseline, which represents optimal performance (direct supervision using high-resolution labels). We use the following standard evaluation metrics: F1 score (the harmonic mean of the precision and recall scores) and mean intersection-over-union (mIOU), also known as the Jaccard Index. For both metrics, we use the {\em macro} strategy which computes the metric for each class separately, and then averages across classes using equal weights for each class.

\subsection{Case Study: Impact of Label Resolution}

To begin, we study the impact of label resolution on prediction performance. For this experiment, we optimize the baselines outlined above, which share the same segmentation architecture as our approach (a modified U-Net with a ResNet-18 backbone). For each baseline, we train for 100 epochs using the Low2High dataset. Model selection is performed using a resolution-specific validation set, treating each label set in isolation.

\tabref{baselines} shows the result of this experiment. As expected, {\em High} performs best, due to being directly supervised using high-resolution ground truth. {\em Low} performs significantly worse, highlighting the performance cost associated with simply upsampling the low-resolution ground truth. Finally, {\em Oracle} performs worst, which represents a model capable of perfectly predicting the low-resolution ground truth. Together, these results show that high-resolution ground-truth is crucial for achieving high quality output in the form of fine-grained predictions.

For comparison, we also show the performance of the label super-resolution approach ({\em Self-Epitomic LSR}) proposed by Malkin et al.\cite{malkin2020mining}, adapting the implementation made available by the authors. This iterative method performs significantly worse than all other methods, despite requiring prior knowledge of statistics that describe how frequently high-resolution annotations occur within low-resolution annotations.

\begin{table}
  \centering
  \caption{Performance of baseline methods.}
  
  \begin{tabular}{@{}lccc@{}}
    \toprule
    & \multicolumn{1}{c}{Acc} & \multicolumn{1}{c}{F1} & \multicolumn{1}{c}{mIOU} \\
    \bottomrule
    Oracle & 80.17\% & 72.69\% & 60.28\% \\
    Low & 84.40\% & 76.16\% & 64.25\% \\
    High & 93.16\% & 90.87\% & 83.59\% \\
    \midrule
    {\em Self-Epitomic LSR}~\cite{malkin2020mining} & 69.11\% & 50.90\% & 37.10\% \\
    \bottomrule
  \end{tabular}
  \label{tbl:baselines}
\end{table}

\begin{table}
  \centering
  \caption{An ablation study highlighting the impact of different components. Our full approach significantly reduces the gap in performance to the high-resolution baseline.}
  
  \begin{tabular}{@{}lccccc@{}}
    \toprule
    & & & \multicolumn{1}{c}{Acc} & \multicolumn{1}{c}{F1} & \multicolumn{1}{c}{mIOU} \\
    \bottomrule
    Agg. & App. & Pre. & & & \\
    \checkmark & & & 82.63\% & 74.59\% & 62.52\% \\
    \checkmark & \checkmark & & 87.43\% & 81.12\% & 69.84\% \\
    \checkmark & \checkmark & \checkmark & \textbf{88.61\%} & \textbf{82.60\%} & \textbf{71.69\%} \\
    \midrule
    \multicolumn{3}{l}{Baseline: High} & 93.16\% & 90.87\% & 83.59\% \\
    \bottomrule
  \end{tabular}
  \label{tbl:ablation}
\end{table}

\begin{figure}
    \centering
    \includegraphics[width=.85\linewidth]{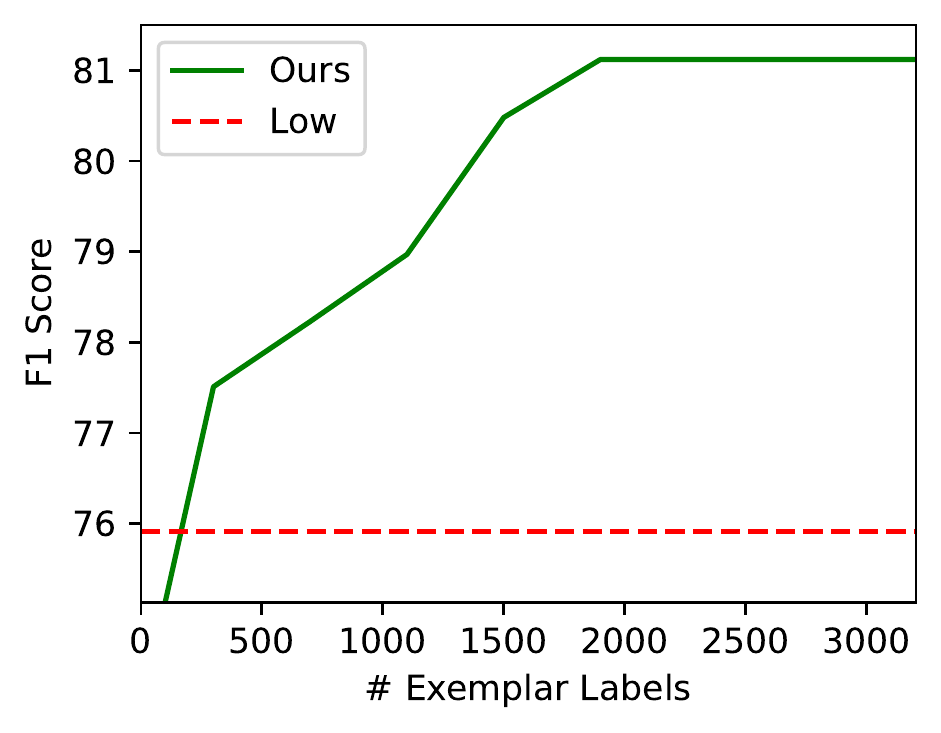}
    \caption{The performance of our proposed approach versus the quantity of exemplar labels used during model training.}
    \label{fig:varying}
\end{figure}

\subsection{Ablation Study}

Next, we evaluate the performance of our method, along with an ablation study showing the performance impact attributed to individual components. \tabref{ablation} shows the outcome of this experiment. The results indicate that using region aggregation (Agg.) alone with direct supervision from low-resolution annotations does not outperform the low-resolution baseline from \tabref{baselines}. This is understandable, as the region aggregation step allows the segmentation network to output fine-grained predictions, but does not constrain the predictions to be realistic. In other words, the aggregation step simply encourages the aggregated prediction to agree with the target label.

\begin{figure*}

  \centering
  
  \setlength\tabcolsep{1pt}
  \newcommand\w{.16\linewidth}    
    
    \begin{tabular}{cccccccc}
    
      Image & GT (High) & Pred (High) & GT (Low) & Pred (Low) & Ours \\
      
      \includegraphics[width=\w]{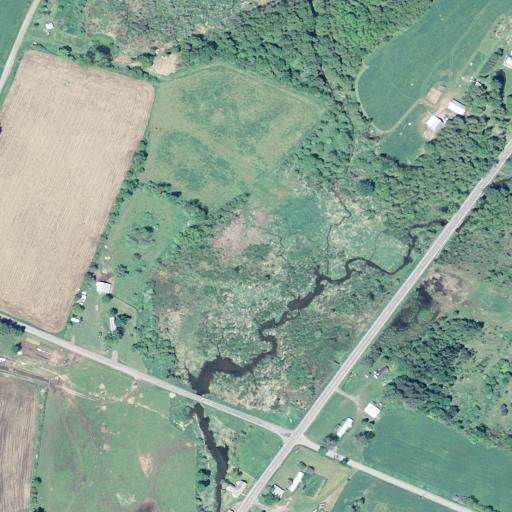} &
      \includegraphics[width=\w]{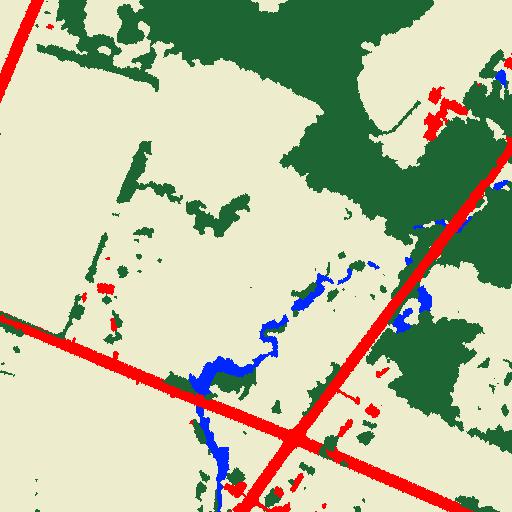} &
      \includegraphics[width=\w]{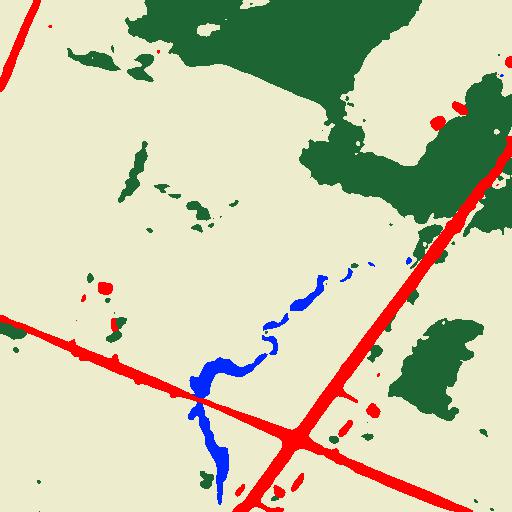} &
      \includegraphics[width=\w]{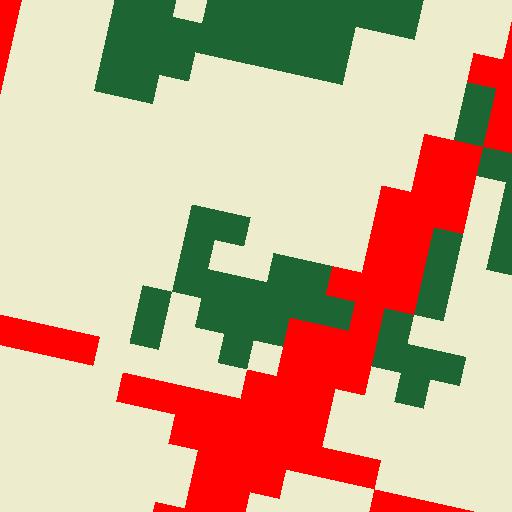} &
      \includegraphics[width=\w]{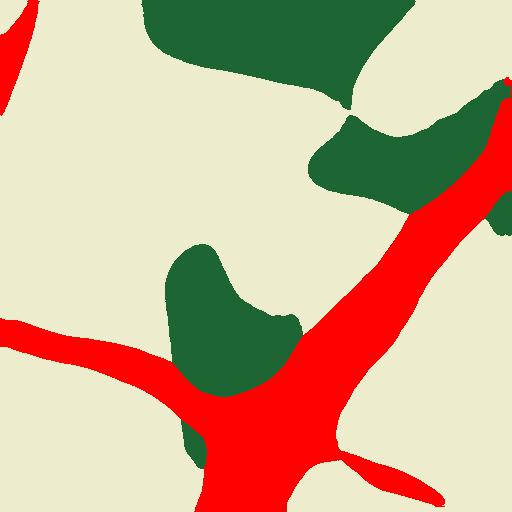} &
      \includegraphics[width=\w]{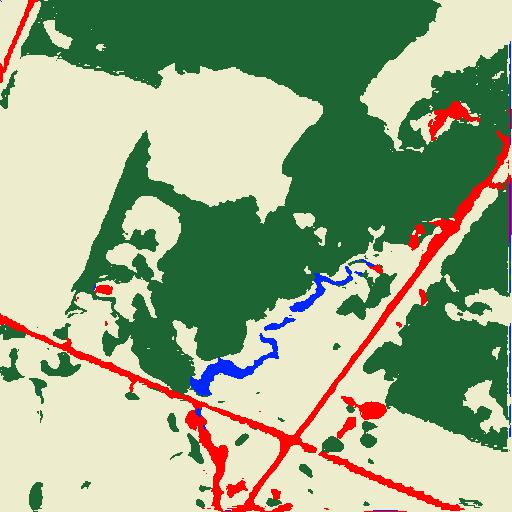} \\
      
      \includegraphics[width=\w]{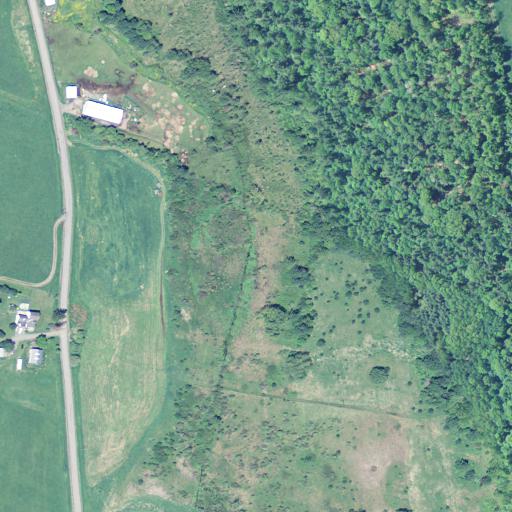} &
      \includegraphics[width=\w]{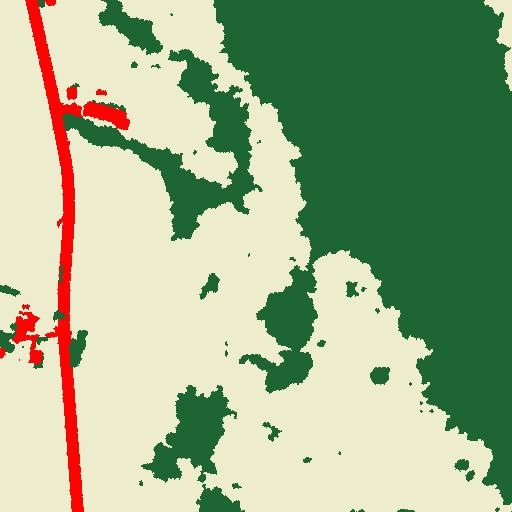} &
      \includegraphics[width=\w]{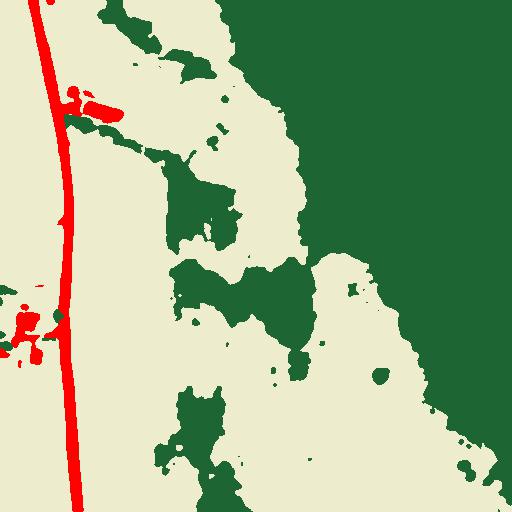} &
      \includegraphics[width=\w]{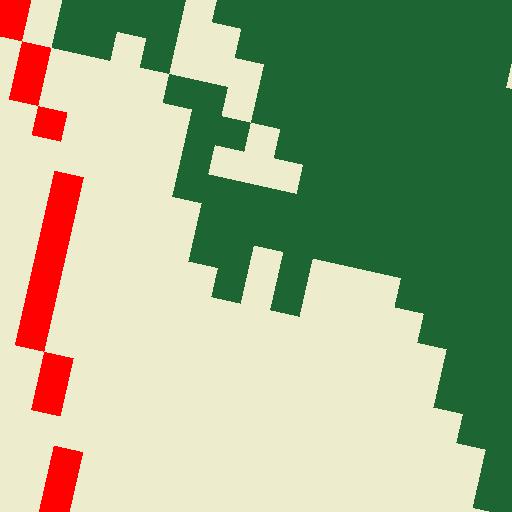} &
      \includegraphics[width=\w]{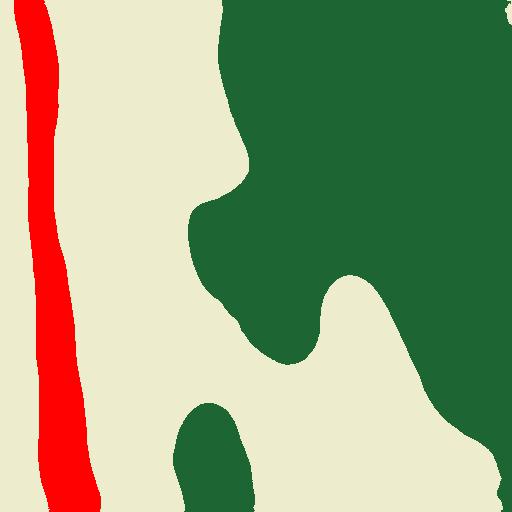} &
      \includegraphics[width=\w]{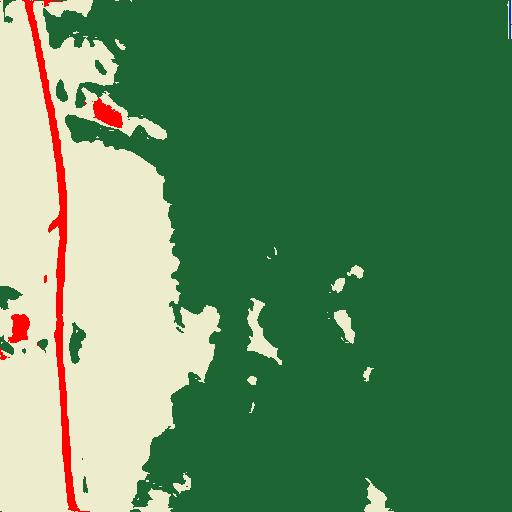} \\
      
      \includegraphics[width=\w]{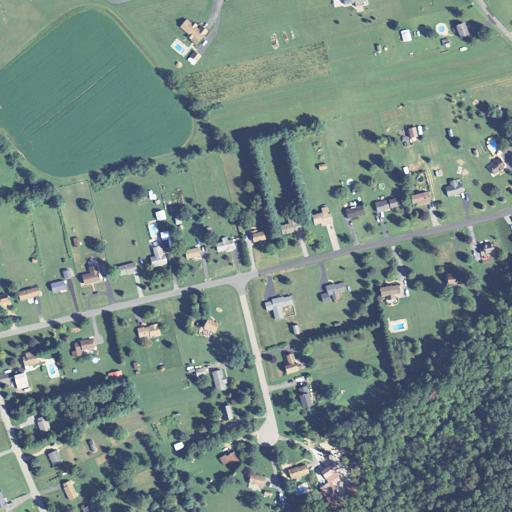} &
      \includegraphics[width=\w]{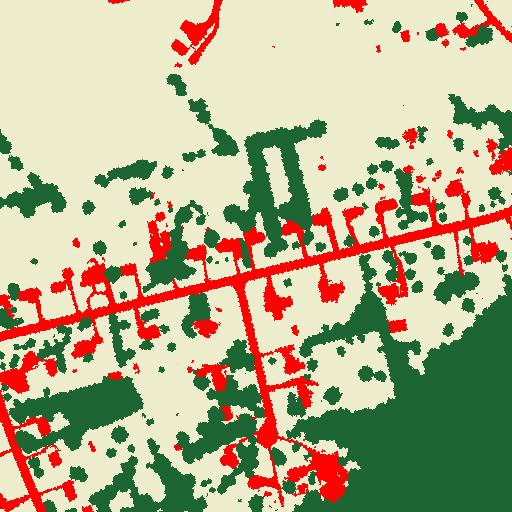} &
      \includegraphics[width=\w]{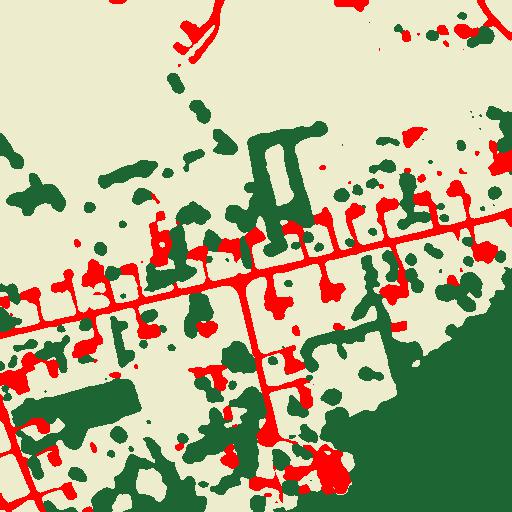} &
      \includegraphics[width=\w]{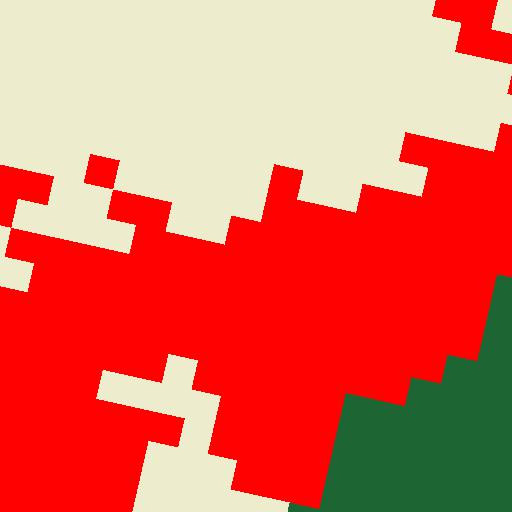} &
      \includegraphics[width=\w]{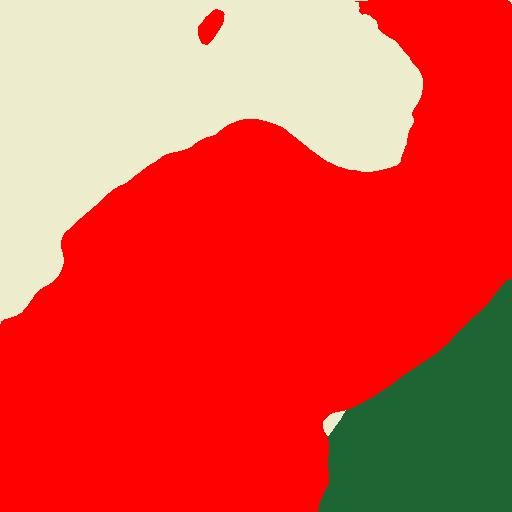} &
      \includegraphics[width=\w]{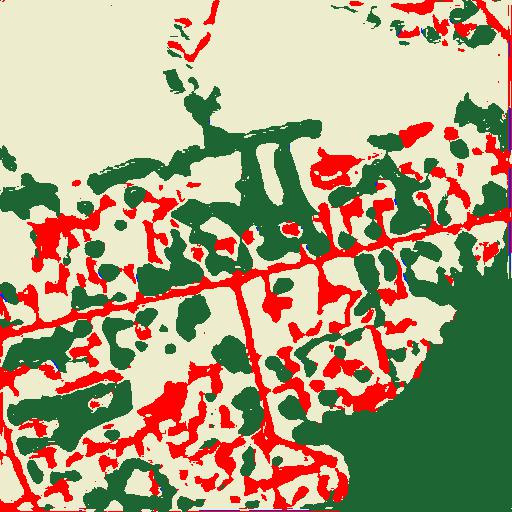} \\
      
      \includegraphics[width=\w]{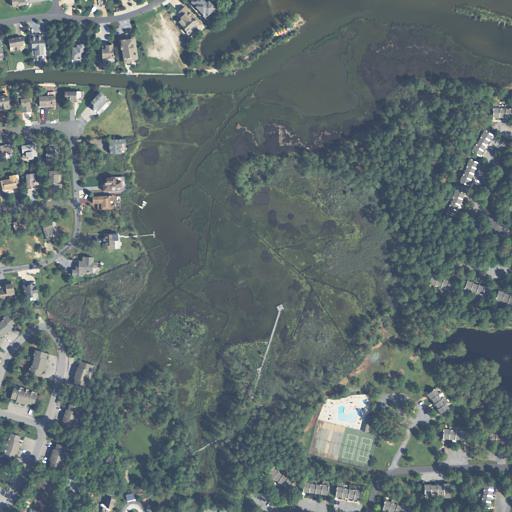} &
      \includegraphics[width=\w]{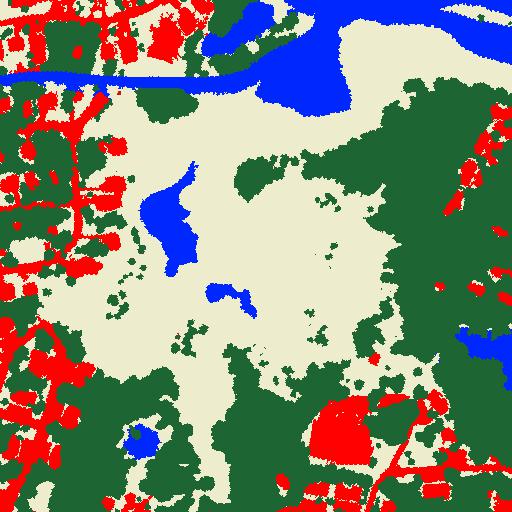} &
      \includegraphics[width=\w]{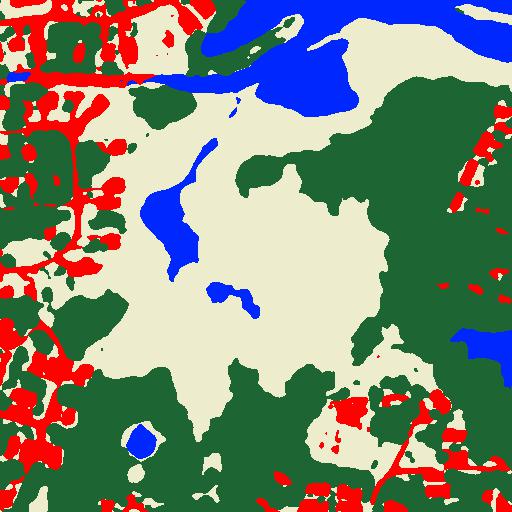} &
      \includegraphics[width=\w]{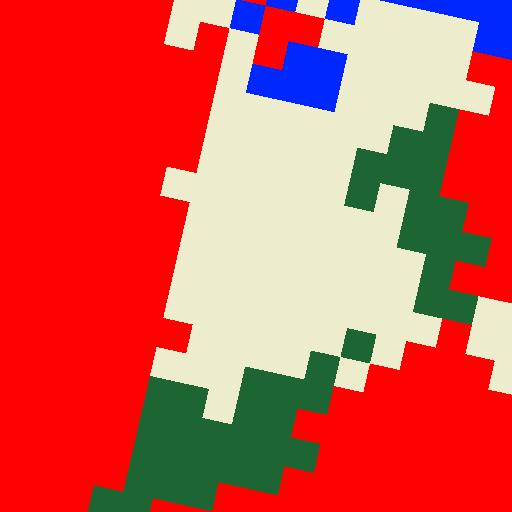} &
      \includegraphics[width=\w]{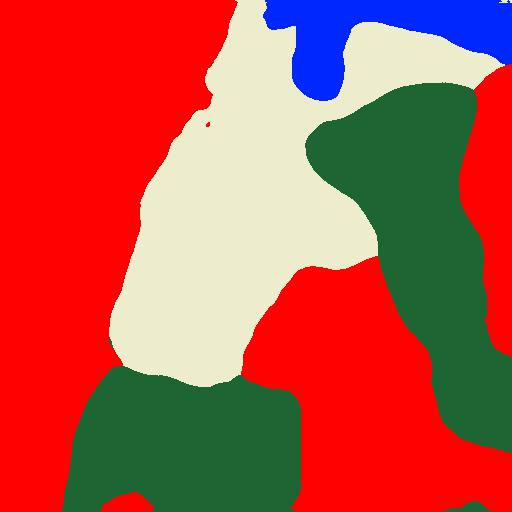} &
      \includegraphics[width=\w]{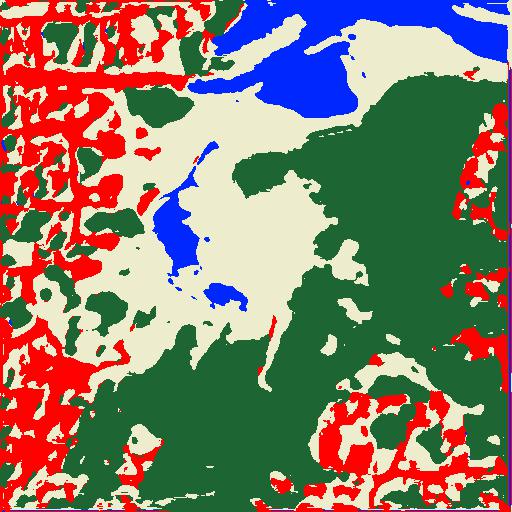} \\
      
      \includegraphics[width=\w]{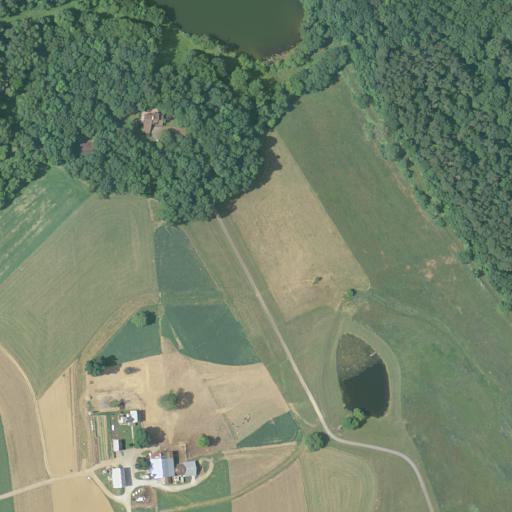} &
      \includegraphics[width=\w]{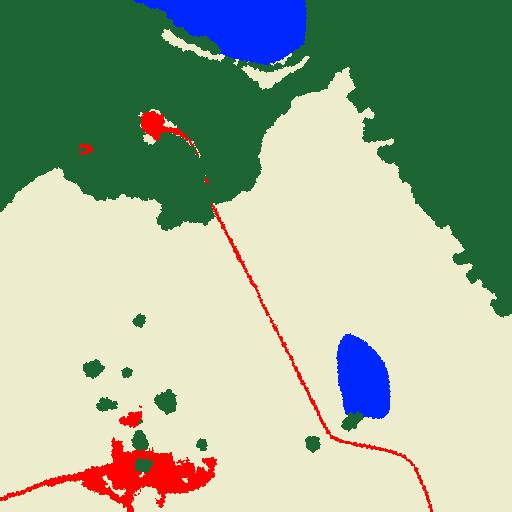} &
      \includegraphics[width=\w]{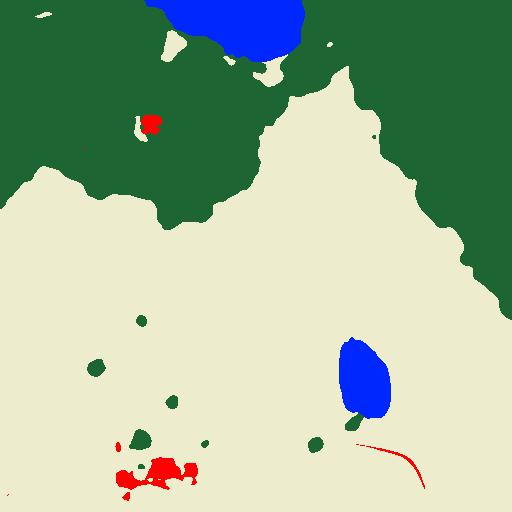} &
      \includegraphics[width=\w]{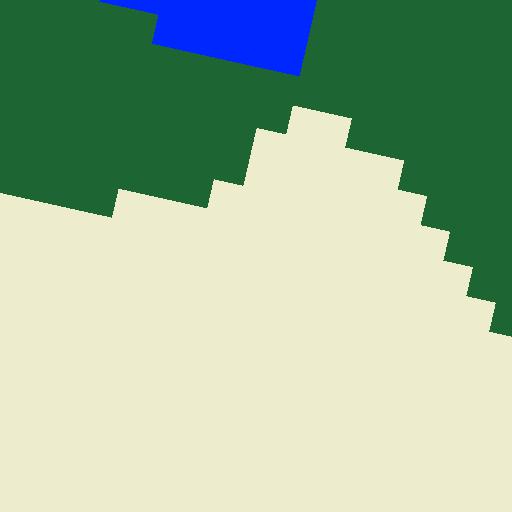} &
      \includegraphics[width=\w]{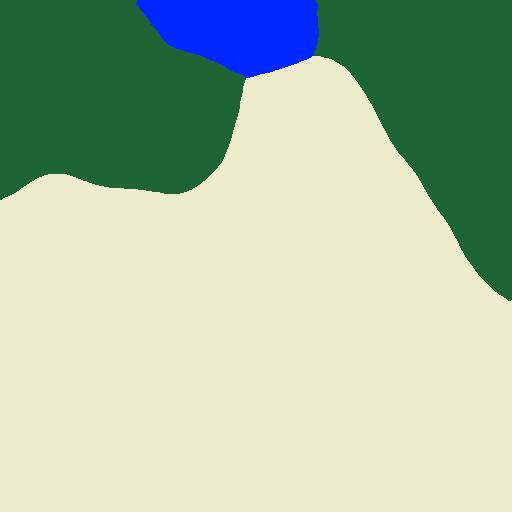} &
      \includegraphics[width=\w]{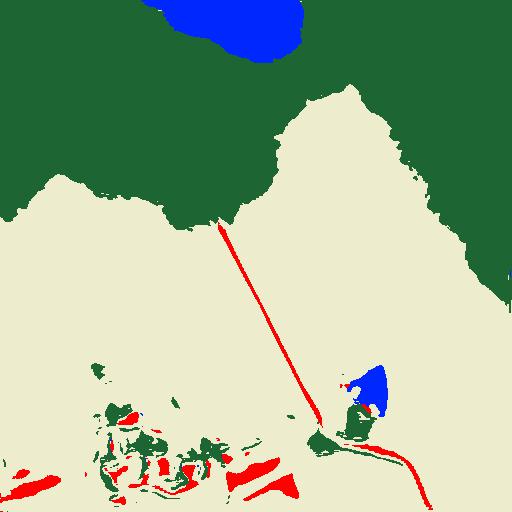} \\
    
    \end{tabular}

  \caption{Example qualitative results. From left to right: input image, high-resolution ground-truth label, prediction from the high-resolution baseline, low-resolution ground-truth label, prediction from the low-resolution baseline, and our result.}

  \label{fig:qualitative_results}

\end{figure*}

However, including appearance-based adversarial training (App.) together with region aggregation (Agg.) results in performance that surpasses both the low-resolution and oracle baselines. This component encourages the output predictions to look similar in appearance to an exemplar set of high-resolution annotations. Finally, including self-supervised pretraining (Pre.) results in a 4.21\% accuracy, 6.44\% F1, and 7.44\% mIOU gain over the low-resolution baseline. The full model's performance approaches the high-resolution baseline, without requiring direct supervision from high-resolution labels. \figref{qualitative_results} shows example outputs from our approach versus baselines.

\subsection{Dependence on Exemplar Labels}

Our method takes advantage of an exemplar set of high-resolution labels to guide the learning process. These labels have no correspondence with the input imagery, but encourage our method to produce realistic (i.e., fine-grained) predictions. Therefore, we evaluate how the performance of our proposed approach changes with respect to the quantity of exemplar labels used during model training. For this experiment, we use our full approach but omit the self-supervised pretraining component.

\figref{varying} visualizes the results of this study. The x-axis represents the number of high-resolution labels used during model training, and the y-axis indicates the resulting model's F1 score on the test set. As expected, model performance increases as the quantity of exemplar labels increases. Even with very few exemplar labels, our method is able to significantly outperform the low-resolution baseline.

\begin{table}
  \centering
  \caption{Domain adaption study with and without self-supervised pretraining (using our full model).}
  
  \begin{tabular}{@{}lccccc@{}}
    \toprule
    \multicolumn{1}{c}{Pretrain} & \multicolumn{1}{c}{Test} & \multicolumn{1}{c}{Acc} & \multicolumn{1}{c}{F1} & \multicolumn{1}{c}{mIOU} \\
    \bottomrule
    None & Ches$^-$ & 86.36\% & 80.07\% & 68.09\% \\
    Ches$^-$ & Ches$^-$ & 88.38\% & 80.31\% & 68.68\% \\
    \textbf{USA*} & Ches$^-$ & \textbf{88.91\%} & \textbf{82.19\%} & \textbf{71.01\%} \\
    \midrule
    None & VA & 85.50\% & 76.03\% & 63.35\% \\
    Ches$^-$ & VA & 86.54\% & 77.03\% & 65.04\% \\
    \textbf{USA*} & VA & \textbf{88.40\%} & \textbf{79.03\%} & \textbf{67.81\%} \\
    \midrule
    None & MWI & 55.93\% & 44.90\% & 30.81\% \\
    Ches$^-$ & MWI & 55.41\% & 45.21\% & 30.99\% \\
    \textbf{USA*} & MWI & \textbf{63.32\%} & \textbf{48.52\%} & \textbf{35.51\%} \\
    \bottomrule
  \end{tabular}
  \label{tbl:domain_adaption}
\end{table}

\subsection{Generalizing to Novel Locations}

We use the Low2High dataset to investigate the impact of self-supervised pretraining on model generalization. For this experiment, we use the MAE image reconstruction strategy outlined in \secref{self-supervised} to pretrain our method, but vary the set of unlabeled images. The first set, Ches$^-$, represents imagery from the training set minus Virginia (VA). VA was excluded to serve as a held-out test region from the Chesapeake Bay area. For the second set, we use the auxiliary imagery from Low2High (\secref{auxiliary}), denoted as USA*. After pretraining for 100 epochs, the resulting models are fine-tuned for 100 epochs on the training portion of Ches$^-$ (states: DE, MD, NY, WV, and PA). Finally, we evaluate each of the two pretraining strategies and a baseline representing no pretraining (None) on three downstream held-out test sets: the test portion of Ches$^-$, test samples from VA, and our new test set over Milwaukee, WI (MWI).

\tabref{domain_adaption} summarizes the results of this experiment. The first three rows show performance for the Ches$^-$ test set. As expected, no pretraining (None) performs the worst. Notably, pretraining on USA* outperforms pretraining on Ches$^-$. As both pretraining sets have exposure to imagery similar to the testing region, this can likely be attributed to learning an improved feature representation from a more geographically diverse set of images.

The middle three rows show performance on the held-out VA test set. Unlike the prior three rows, no method had any previous exposure to VA imagery during model training. Despite this, the results follow the same trend as before. No pretraining (None) performs the worst and pretraining on USA* outperforms pretraining on Ches$^-$. This result shows that pretraining on a more diverse set of images (USA*) is still beneficial for downstream learning, as compared to pretraining only on a set of images (Ches$^-$) that is more geographically similar to the test set.

Finally, the bottom three rows show performance for the held-out MWI test set. In this scenario, the baseline (None) is competitive with Ches$^-$. This is understandable, as MWI is geographically diverse compared to the Chesapeake Bay region. USA* outperforms both alternatives, showcasing that self-supervised pretraining has a significant positive impact on generalizing a model to novel locations.

\section{Conclusion}

Semantic segmentation has reached impressive performance levels, but in remote sensing, it is still extremely challenging to acquire high-resolution annotations. In practice, for many tasks it is much easier to get coarse, low-resolution labels. Our results demonstrate how naively upsampling these labels results in low quality outputs that lack sufficient detail. Instead, we proposed a method that is supervised using low-resolution annotations, incorporates exemplar high-resolution labels, and generates fine-grained output. Experiments on a novel dataset demonstrate how our approach significantly reduces the performance gap to a high-resolution baseline. This approach has the potential to have real-world applicability in the remote sensing domain.

\ifwacvfinal
{
\normalsize
\paragraph*{Acknowledgments}
This research is based upon work supported in part by the Office of the Director of National Intelligence (ODNI), Intelligence Advanced Research Projects Activity (IARPA), via  2021-2011000005. The views and conclusions contained herein are those of the authors and should not be interpreted as necessarily representing the official policies, either expressed or implied, of ODNI, IARPA, or the U.S. Government. The U.S. Government is authorized to reproduce and distribute reprints for governmental purposes notwithstanding any copyright annotation therein.
}
\fi

{\small
\bibliographystyle{ieee_fullname}
\bibliography{biblio}
}

\null
\vskip .375in
\twocolumn[{%
  \begin{center}
    \textbf{\Large Supplemental Material : \\ Handling Image and Label Resolution Mismatch in Remote Sensing}
  \end{center}
  \vspace*{24pt}
}]
\setcounter{section}{0}
\setcounter{equation}{0}
\setcounter{figure}{0}
\setcounter{table}{0}
\makeatletter
\renewcommand{\theequation}{S\arabic{equation}}
\renewcommand{\thefigure}{S\arabic{figure}}
\renewcommand{\thetable}{S\arabic{table}}

This document contains additional details and experiments related to our methods.

\section{Dataset Details}

\begin{figure}
    \centering
    \includegraphics[width=1\linewidth]{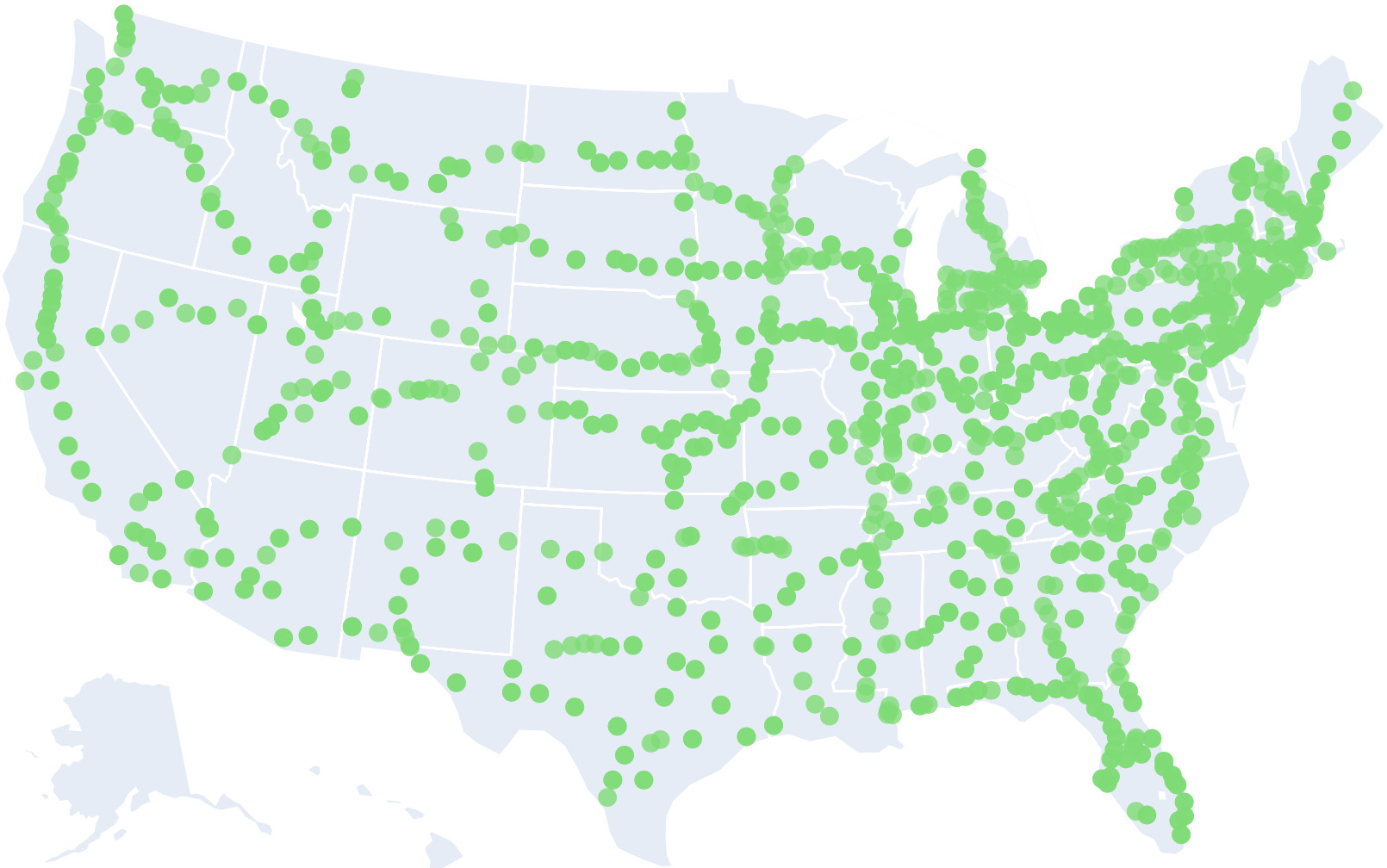}
    \caption{Spatial coverage of the auxiliary imagery in the Low2High dataset.}
    \label{fig:coverage}
\end{figure}

\begin{figure}

  \centering
  
  \setlength\tabcolsep{1pt}
  \newcommand\w{.24\linewidth}    
    
    \begin{tabular}{cccc}
    
      \includegraphics[width=\w]{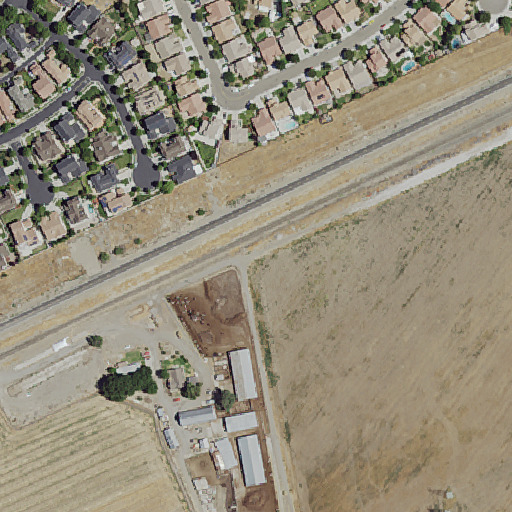} &
      \includegraphics[width=\w]{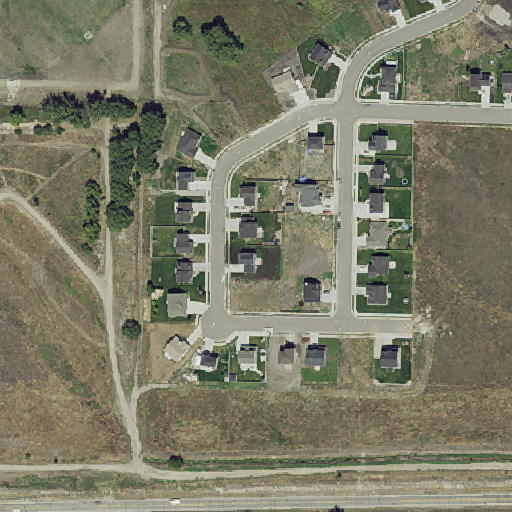} &
      \includegraphics[width=\w]{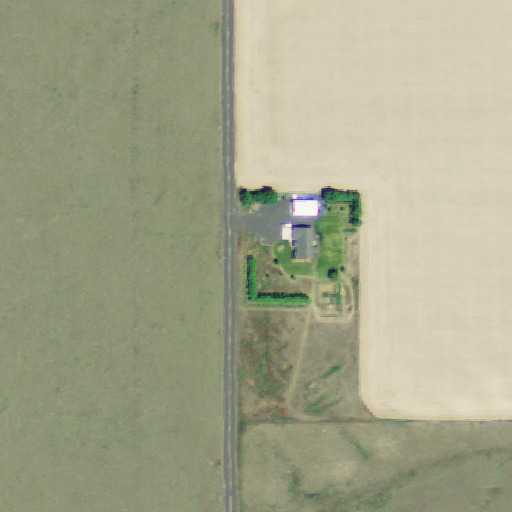} &
      \includegraphics[width=\w]{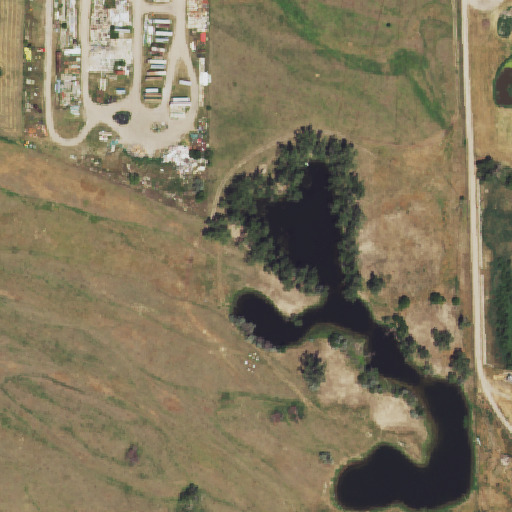} \\
      
      \includegraphics[width=\w]{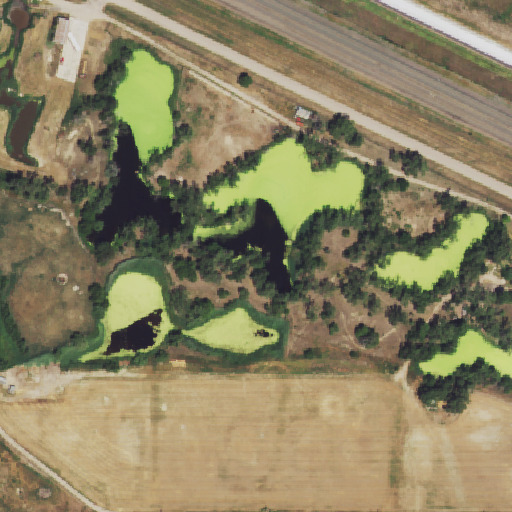} &
      \includegraphics[width=\w]{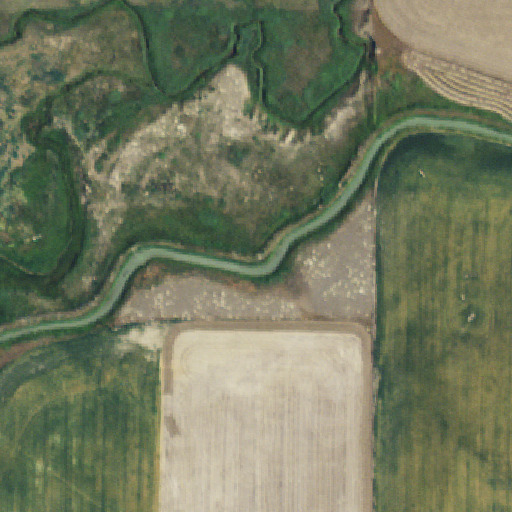} &
      \includegraphics[width=\w]{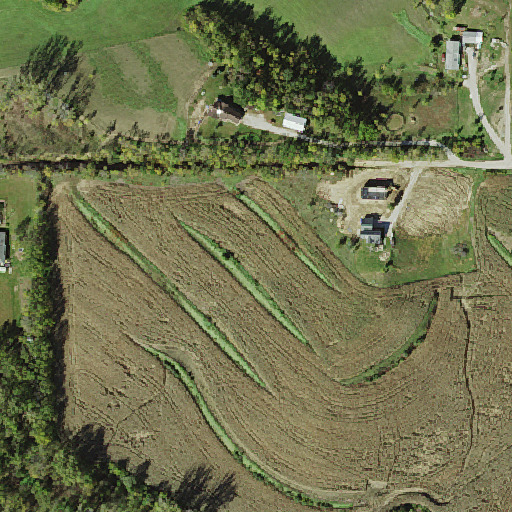} &
      \includegraphics[width=\w]{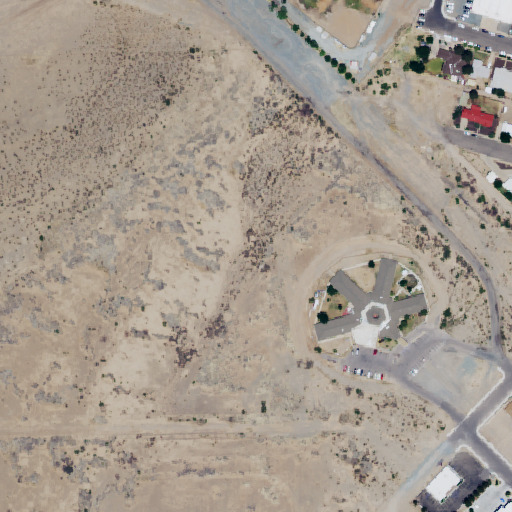} \\
    
    \end{tabular}

  \caption{Examples of the auxiliary imagery in the Low2High dataset.}

  \label{fig:usa}

\end{figure}

We introduced the Low2High dataset, an extension of the Chesapeake~\cite{robinson2019large} dataset that includes a merged label taxonomy, additional auxiliary imagery across the United States, and a new held-out test set from Milwaukee, WI. The spatial coverage of the auxiliary imagery is shown in \figref{coverage}. Example data from the held-out test set is shown in \figref{usa}. The merged label taxonomy we use for our experiments is outlined in \tabref{taxonomy}. A summary of the number of samples per subset of the primary and auxiliary dataset components are included in \tabref{dataset_details} and \tabref{aux_dataset_details}, respectively.

\section{Self-Supervised Learning}

\begin{figure*}
    \centering
    \includegraphics[trim={5px 0 5px 0},clip,width=.8\linewidth]{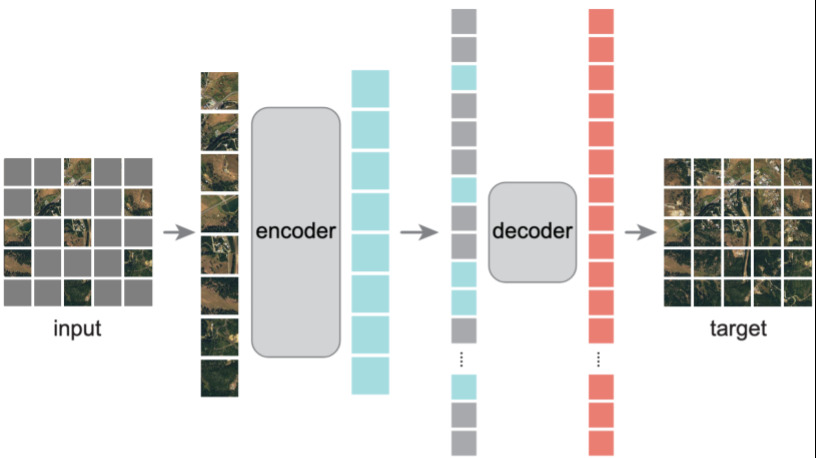}
    \caption{An overview of the masked autoencoders architecture.}
    \label{fig:mae_architecture}
\end{figure*}

\begin{figure*}

  \centering
  
  \setlength\tabcolsep{1pt}
  \newcommand\w{.24\linewidth}    
    
    \begin{tabular}{cccc}
    
      Image & Masked & Reconstruction & Composite \\
    
      \includegraphics[width=\w]{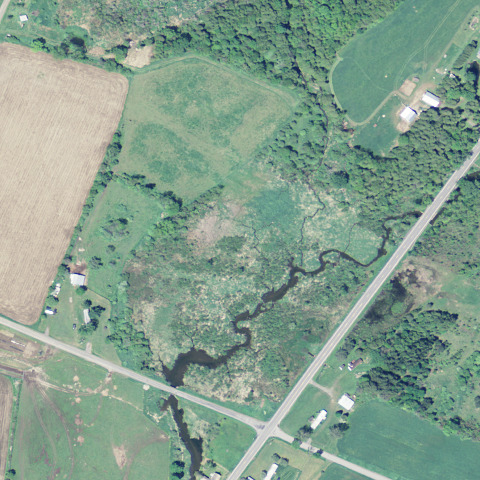} &
      \includegraphics[width=\w]{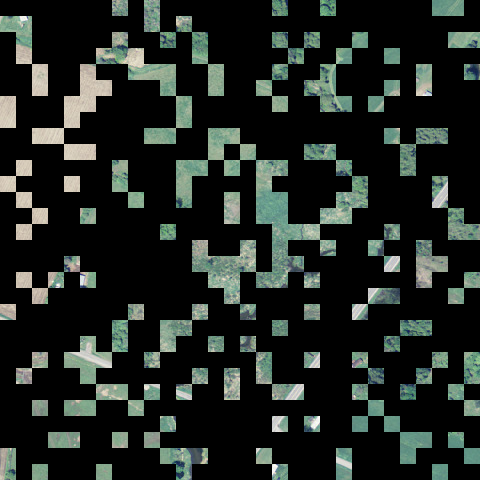} &
      \includegraphics[width=\w]{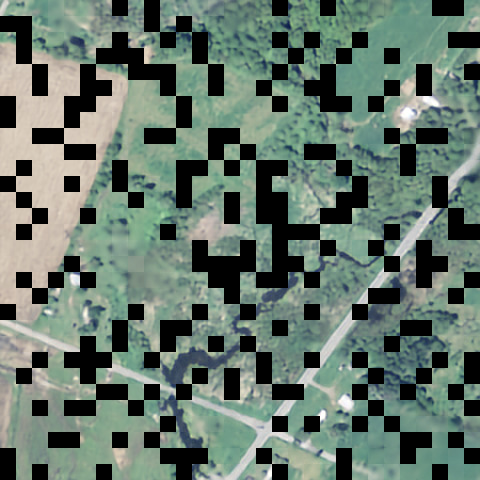} &
      \includegraphics[width=\w]{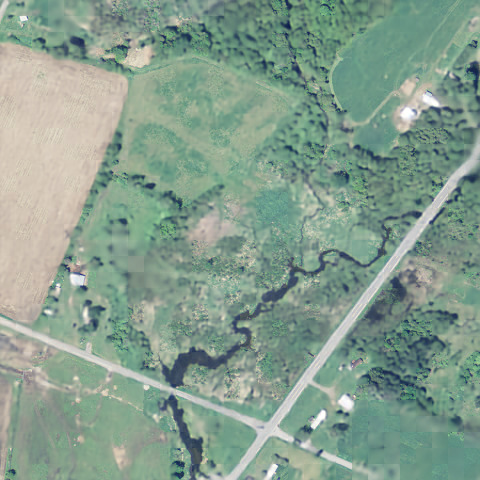}\\
    
    \end{tabular}

  \caption{An example reconstruction from masked autoencoders on overhead imagery.}

  \label{fig:reconstruction}

\end{figure*}

We show that self-supervised pretraining on the auxiliary image dataset in the Low2High dataset leads to improved performance when evaluating on regions with drastically different appearance characteristics. We use image reconstruction as a pretraining strategy using the masked autoencoders (MAE)~\cite{he2022masked} framework, depicted in \figref{mae_architecture}. We used mean squared error between the original image ($I$) and reconstructed image ($R$) for a given batch $B$ as the objective function, as follows:
\begin{equation}
    \frac{1}{B} \sum_i^B (I[i] - R[i])^2
\end{equation}
In \figref{reconstruction} we show a qualitative example of reconstruction for a given overhead image from our test set. The resulting composite (masked region combined with reconstruction) image appears to be a near duplicate of the input image after training for 300 epochs.

\begin{table}
  \centering
  \caption{Our merged label taxonomy.}
  
  \resizebox{.9\linewidth}{!}{
      \begin{tabular}{@{}ll@{}}
        \toprule
        Chesapeake~\cite{robinson2019large} & Remapped \\
        \midrule
        Water & Water \\
        Forest & Forest \\
        Field & Field \\
        Barren & Impervious \\
        Impervious (other) & Impervious \\
        Impervious (road) & Impervious \\ 
        \midrule
        NLCD~\cite{yang2018new} & \\
        \midrule
        Open Water & Water \\
        Deciduous Forest & Forest \\
        Evergreen Forest & Forest \\
        Mixed Forest & Forest \\
        Dwarf Scrub & Forest \\
        Shrub/Scrub & Forest \\
        Woody Wetlands & Forest \\
        Grassland/Herbaceous & Field \\
        Sedge/Herbaceous & Field \\
        Lichens & Field \\
        Moss & Field \\
        Pasture/Hay & Field \\
        Cultivated Crops & Field\\
        Emergent Herbaceous Wetlands & Field \\
        Developed, Open Space & Impervious \\
        Developed, Low Intensity & Impervious \\
        Developed, Medium Intensity & Impervious\\
        Developed, High Intensity & Impervious \\
        Barren Land (Rock/Sand/Clay) & Impervious \\
        Perennial Ice/Snow & {\em Ignore} \\
        \midrule
        EPA EnviroAtlas~\cite{pilant2020us} & \\
        \midrule
        Water & Water \\
        Tree & Forest \\
        Shrub & Forest \\
        Orchards & Forest \\
        Wooded Wetlands & Forest \\
        Soil & Field \\
        Grass & Field \\
        Agriculture & Field \\
        Wetlands & Field \\
        Impervious & Impervious \\
        \bottomrule
      \end{tabular}
  }
  
  \label{tbl:taxonomy}
\end{table}

\section{Additional Results}

\begin{figure*}

  \centering
  
  \setlength\tabcolsep{1pt}
  \newcommand\w{.19\linewidth}    
    
    \begin{tabular}{ccccc}
    
      Image & GT (High) & GT (Low) & Ours (Agg.) & Ours \\
    
      \includegraphics[width=\w]{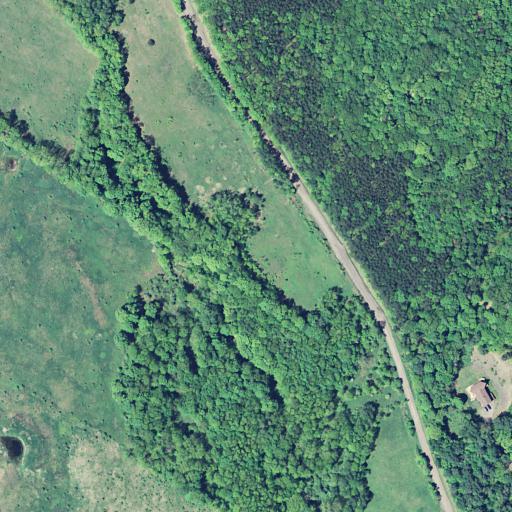} &
      \includegraphics[width=\w]{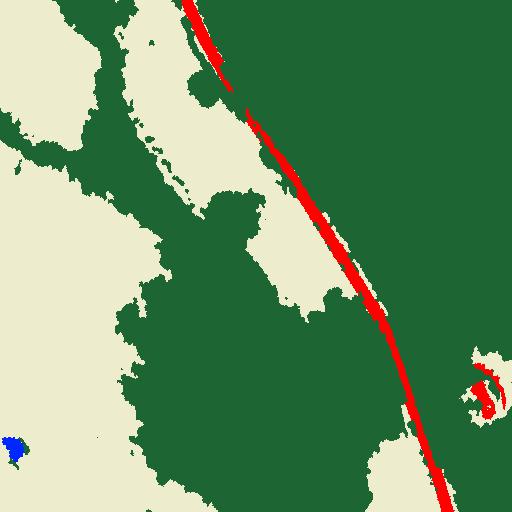} &
      \includegraphics[width=\w]{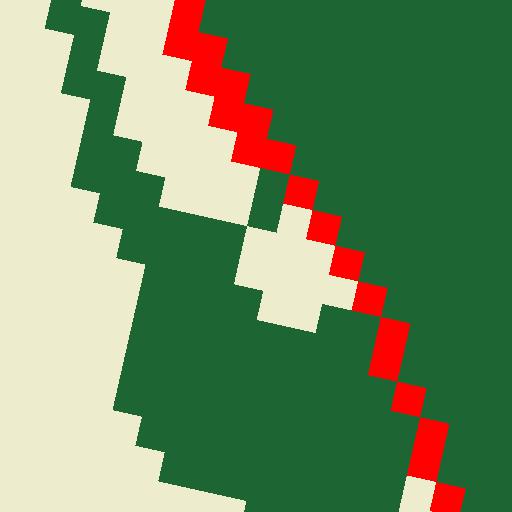} &
      \includegraphics[width=\w]{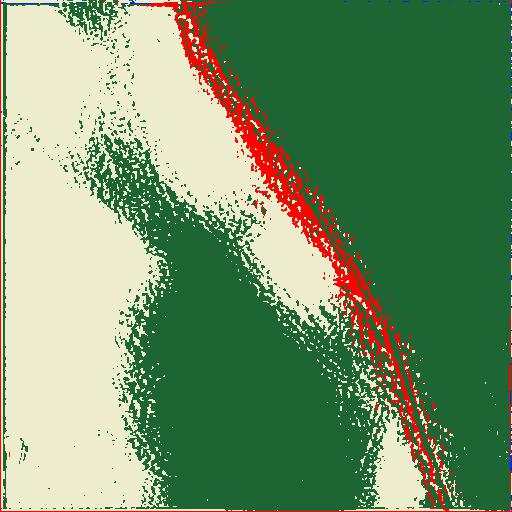} &
      \includegraphics[width=\w]{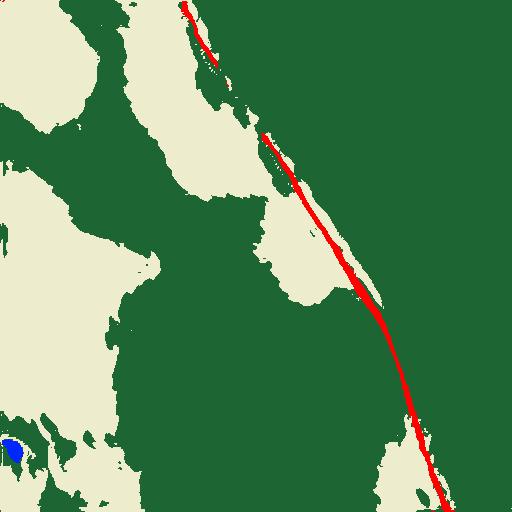} \\
      
      \includegraphics[width=\w]{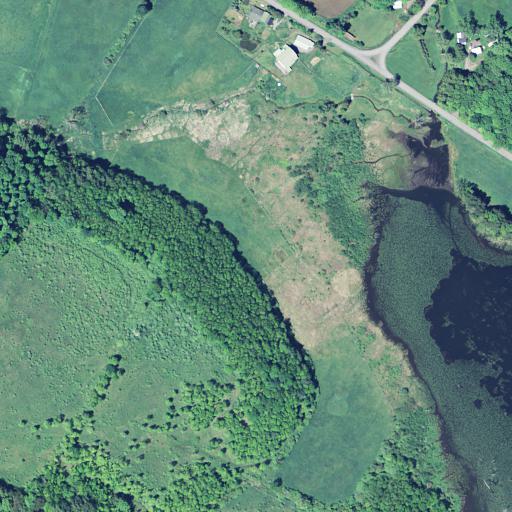} &
      \includegraphics[width=\w]{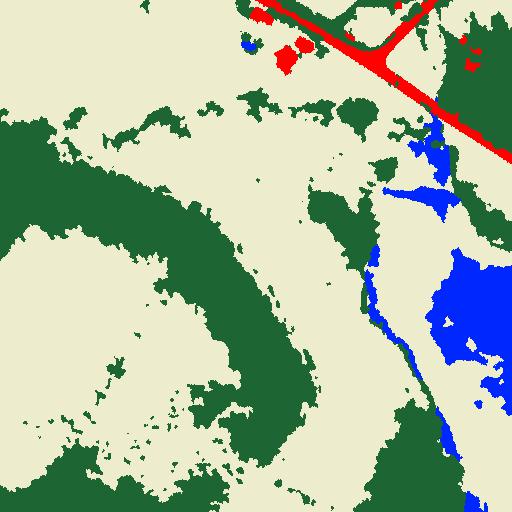} &
      \includegraphics[width=\w]{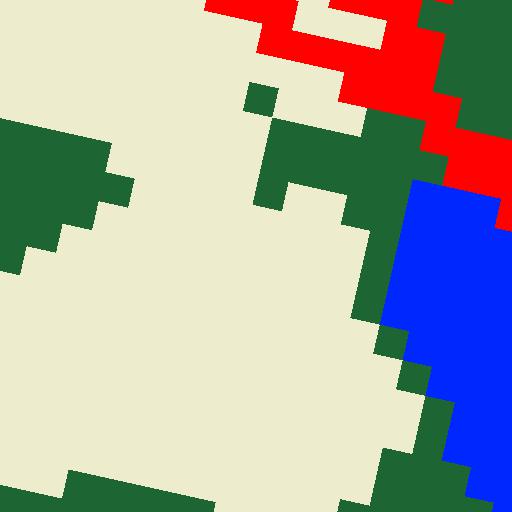} &
      \includegraphics[width=\w]{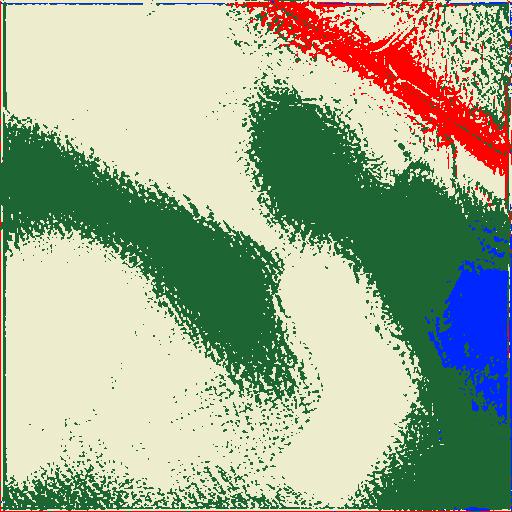} &
      \includegraphics[width=\w]{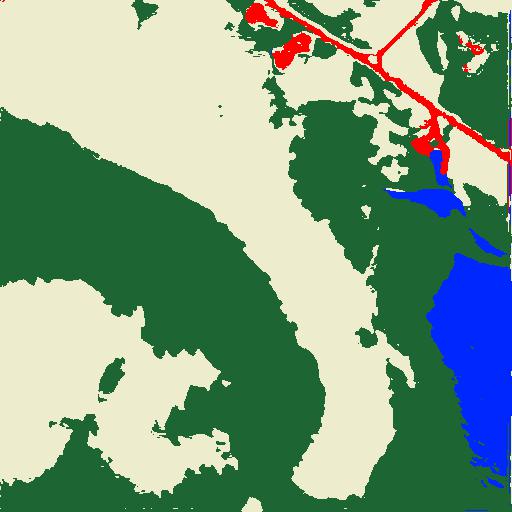} \\
    
    \end{tabular}

  \caption{Example qualitative results from our approach.}

  \label{fig:aggregation}

\end{figure*}

\begin{figure*}

  \centering
  
  \setlength\tabcolsep{1pt}
  \newcommand\w{.19\linewidth}    
    
    \begin{tabular}{ccccc}
    
      \includegraphics[width=\w]{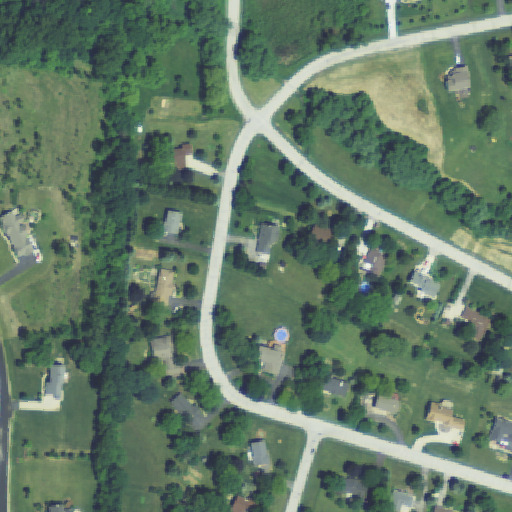} &
      \includegraphics[width=\w]{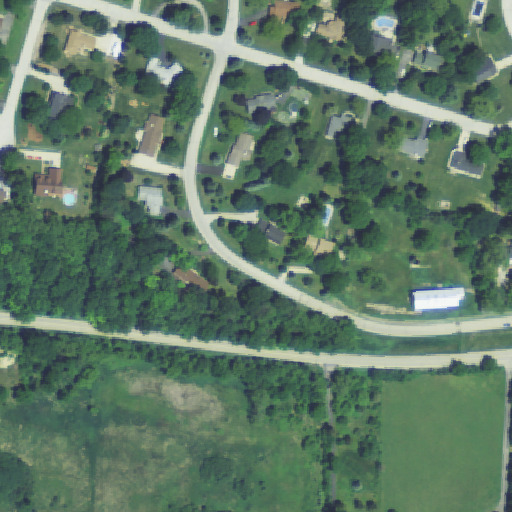} &
      \includegraphics[width=\w]{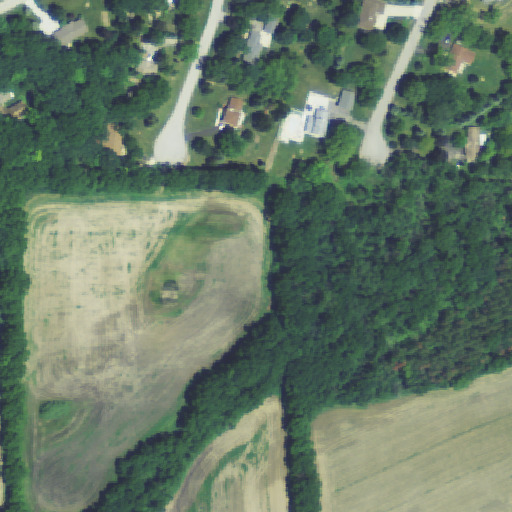} &
      \includegraphics[width=\w]{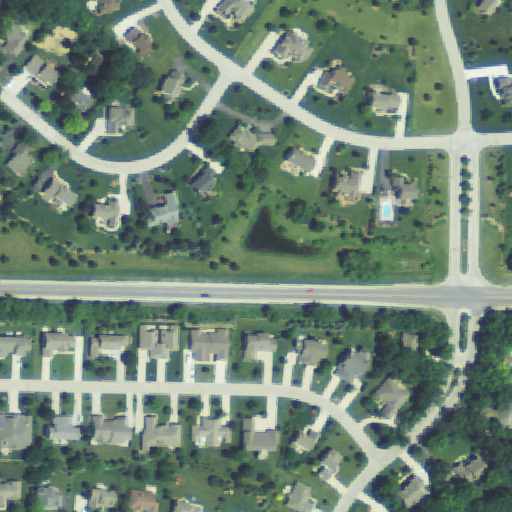} &
      \includegraphics[width=\w]{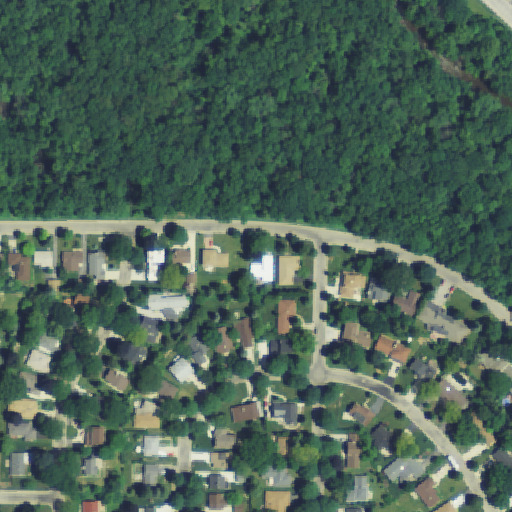} \\
      
      \includegraphics[width=\w]{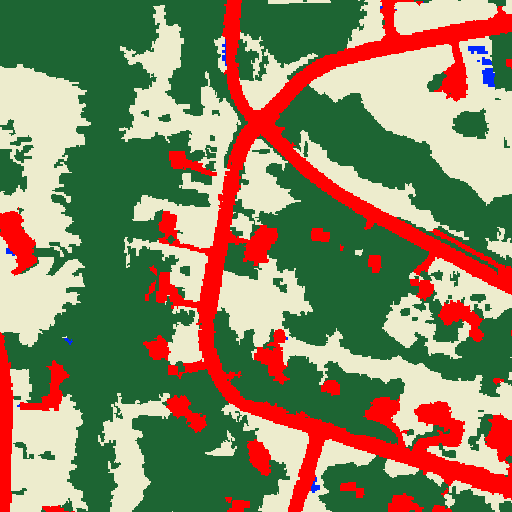} &
      \includegraphics[width=\w]{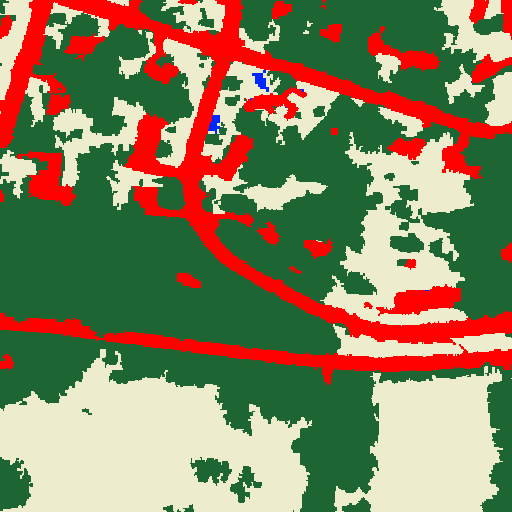} &
      \includegraphics[width=\w]{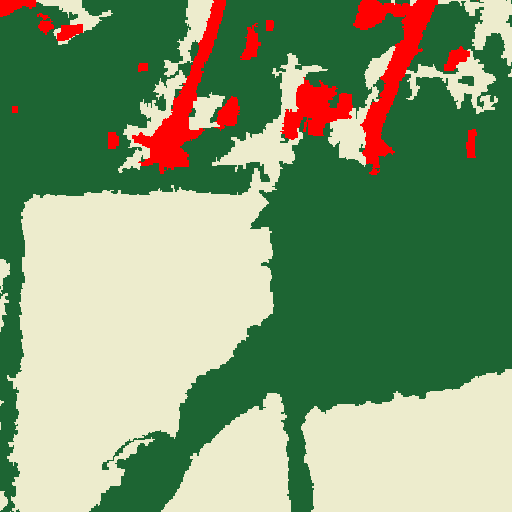} &
      \includegraphics[width=\w]{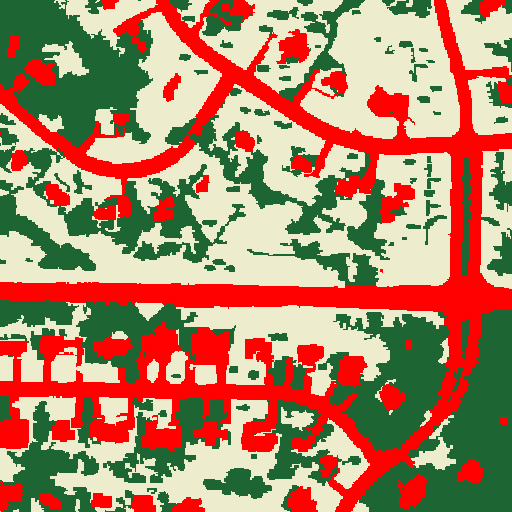} &
      \includegraphics[width=\w]{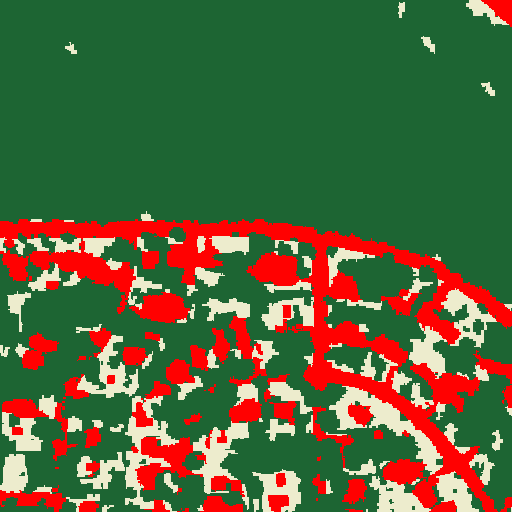} \\

      \includegraphics[width=\w]{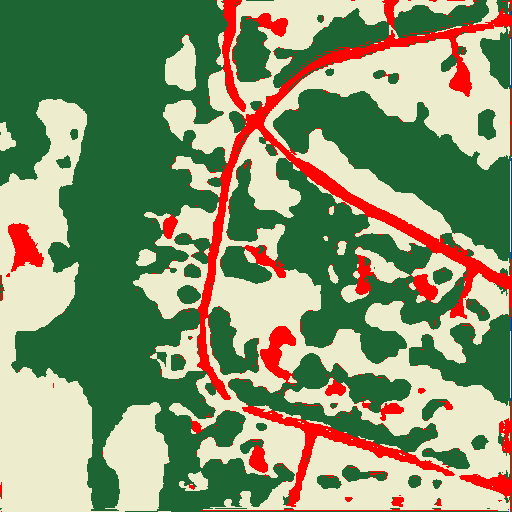} &
      \includegraphics[width=\w]{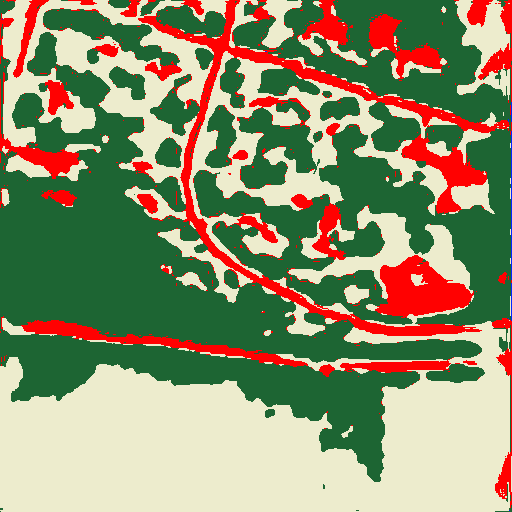} &
      \includegraphics[width=\w]{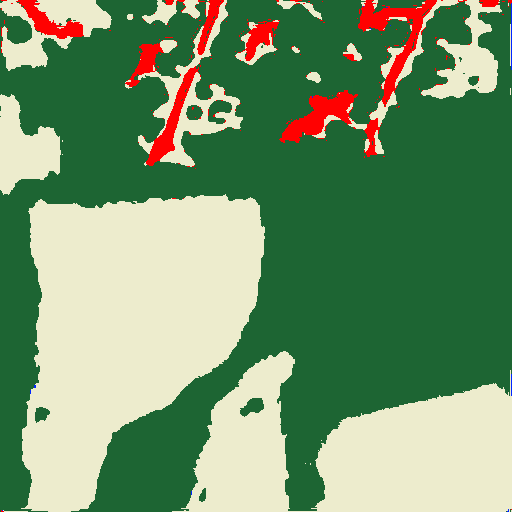} &
      \includegraphics[width=\w]{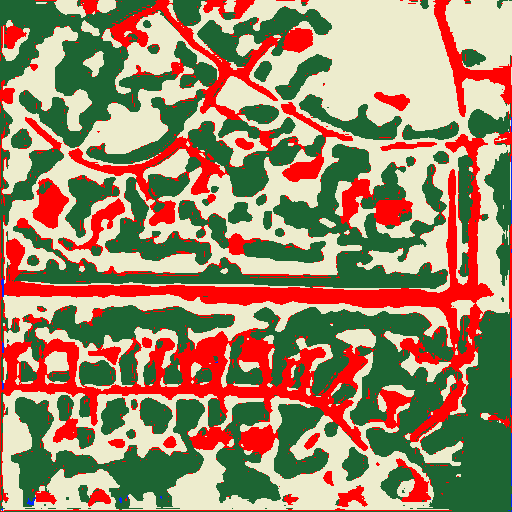} &
      \includegraphics[width=\w]{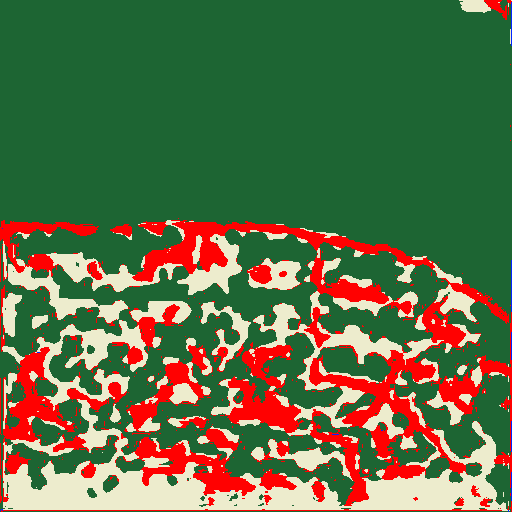} \\
    
    \end{tabular}

  \caption{Example qualitative results for the Milwaukee, WI test set. (top) NAIP image, (middle) remapped ground-truth label from EnviroAtlas, and (bottom) our prediction.}

  \label{fig:mwi_results}

\end{figure*}

\figref{aggregation} shows example qualitative results from the region aggregation component of our method. Region aggregation allows the network to output fine-grained predictions, without requiring the target label to be upsampled to the native image resolution. In other words, the network outputs high-resolution predictions which are aggregated (by summing the logits of pixels in the region) and passed to the objective function for comparison with the low-resolution target label. As observed, this process is often not sufficient by itself due to the low quality of the target label (NLCD is 30m per pixel), as well as discrepancies with the high-resolution label (i.e., missing objects). However, our full method (\figref{aggregation}, right) is able to leverage region aggregation and adversarial learning to yield high quality predictions. \figref{mwi_results} shows qualitative results from the Milwaukee, WI test set, a diverse geographic region not observed during training.

\begin{table}
  \centering
  \caption{Low2High dataset summary (images of size $512 \times 512$).}
  
  \resizebox{1\linewidth}{!}{
      \begin{tabular}{@{}lcccc@{}}
        \toprule
         & Train & Validation & Held-out & Test \\
        \bottomrule
        Delaware      & 4560 & 571 & 571 & 540 \\
        New York      & 3915 & 490 & 490 & 492 \\
        Maryland      & 4886 & 611 & 611 & 540 \\
        Pennsylvania  & 4433 & 555 & 555 & 540 \\
        Virginia      & 4440 & 556 & 556 & 552 \\
        West Virginia & 3029 & 379 & 379 & 540 \\
        \midrule
        Total         & 25263 & 3162 & 3162 & 3204 \\
        \bottomrule
      \end{tabular}
  }
  
  \label{tbl:dataset_details}
\end{table}

\begin{table}
  \centering
  \caption{Variant of the Low2High dataset, including auxiliary imagery, that is used for our model generalization experiments. Rest of USA excludes Alaska, Rhode Island, Hawaii and the other states listed below.}
  
  \resizebox{1\linewidth}{!}{
      \begin{tabular}{@{}lccccc@{}}
        \toprule
        & Auxiliary & Train & Validation & Held-out & Test \\
        \midrule
        Delaware      & 18 & 4560 & 571 & 571 & 540 \\
        New York      & 558 & 3915 & 490 & 490 & 492 \\
        Maryland      & 117 & 4886 & 611 & 611 & 540 \\
        Pennsylvania  & 630 & 4433 & 555 & 555 & 540 \\
        Virginia      &  &  & 1159 &  & 552 \\
        West Virginia & 243 & 3029 & 379 & 379 & 540\\
        Wisconsin     &  &  & 270 &  & 3262 \\
        Rest of USA   & 10458 &  &  &  &  \\
        \midrule
        Total         & 12024 & 20823 & 4035 & 2606 & 6466 \\
        \bottomrule
      \end{tabular}
  }
  
  \label{tbl:aux_dataset_details}
\end{table}

\section{Detailed Architecture}

We provide detailed architecture descriptions for the components of our network. \tabref{arch_encoder} shows the feature encoder used for the overhead imagery (ResNet-18). \tabref{arch_decoder} shows our U-Net style decoder used for generating the segmentation output. Finally, \tabref{arch_discriminator} shows the discriminator architecture.

\begin{table*}
  \centering
  \caption{Feature encoder architecture.}
  
  \resizebox{.59\linewidth}{!}{
      \begin{tabular}{@{}lcccc@{}}
        \toprule
        Layer (type:depth-idx) & Input Shape & Kernel Shape & Output Shape & Param \# \\
        \bottomrule
        Encoder & [1, 3, 512, 512] & -- & [1, 64, 128, 128] & -- \\
        — Conv2d: 1-1 & [1, 3, 512, 512] & [7, 7] & [1, 64, 256, 256] & 9,408 \\
        — BatchNorm2d: 1-2 & [1, 64, 256, 256] & -- & [1, 64, 256, 256] & 128 \\
        — ReLU: 1-3 & [1, 64, 256, 256] & -- & [1, 64, 256, 256] & -- \\
        — MaxPool2d: 1-4 & [1, 64, 256, 256] & 3 & [1, 64, 128, 128] & -- \\
        — Sequential: 1-5 & [1, 64, 128, 128] & -- & [1, 64, 128, 128] & -- \\
        —    — BasicBlock: 2-1 & [1, 64, 128, 128] & -- & [1, 64, 128, 128] & -- \\
        —    —    — Conv2d: 3-1 & [1, 64, 128, 128] & [3, 3] & [1, 64, 128, 128] & 36,864 \\
        —    —    — BatchNorm2d: 3-2 & [1, 64, 128, 128] & -- & [1, 64, 128, 128] & 128 \\
        —    —    — ReLU: 3-3 & [1, 64, 128, 128] & -- & [1, 64, 128, 128] & -- \\
        —    —    — Conv2d: 3-4 & [1, 64, 128, 128] & [3, 3] & [1, 64, 128, 128] & 36,864 \\
        —    —    — BatchNorm2d: 3-5 & [1, 64, 128, 128] & -- & [1, 64, 128, 128] & 128 \\
        —    —    — ReLU: 3-6 & [1, 64, 128, 128] & -- & [1, 64, 128, 128] & -- \\
        —    — BasicBlock: 2-2 & [1, 64, 128, 128] & -- & [1, 64, 128, 128] & -- \\
        —    —    — Conv2d: 3-7 & [1, 64, 128, 128] & [3, 3] & [1, 64, 128, 128] & 36,864 \\
        —    —    — BatchNorm2d: 3-8 & [1, 64, 128, 128] & -- & [1, 64, 128, 128] & 128 \\
        —    —    — ReLU: 3-9 & [1, 64, 128, 128] & -- & [1, 64, 128, 128] & -- \\
        —    —    — Conv2d: 3-10 & [1, 64, 128, 128] & [3, 3] & [1, 64, 128, 128] & 36,864 \\
        —    —    — BatchNorm2d: 3-11 & [1, 64, 128, 128] & -- & [1, 64, 128, 128] & 128 \\
        —    —    — ReLU: 3-12 & [1, 64, 128, 128] & -- & [1, 64, 128, 128] & -- \\
        — Sequential: 1-6 & [1, 64, 128, 128] & -- & [1, 128, 64, 64] & -- \\
        —    — BasicBlock: 2-3 & [1, 64, 128, 128] & -- & [1, 128, 64, 64] & -- \\
        —    —    — Conv2d: 3-13 & [1, 64, 128, 128] & [3, 3] & [1, 128, 64, 64] & 73,728 \\
        —    —    — BatchNorm2d: 3-14 & [1, 128, 64, 64] & -- & [1, 128, 64, 64] & 256 \\
        —    —    — ReLU: 3-15 & [1, 128, 64, 64] & -- & [1, 128, 64, 64] & -- \\
        —    —    — Conv2d: 3-16 & [1, 128, 64, 64] & [3, 3] & [1, 128, 64, 64] & 147,456 \\
        —    —    — BatchNorm2d: 3-17 & [1, 128, 64, 64] & -- & [1, 128, 64, 64] & 256 \\
        —    —    — Sequential: 3-18 & [1, 64, 128, 128] & -- & [1, 128, 64, 64] & 8,448 \\
        —    —    — ReLU: 3-19 & [1, 128, 64, 64] & -- & [1, 128, 64, 64] & -- \\
        —    — BasicBlock: 2-4 & [1, 128, 64, 64] & -- & [1, 128, 64, 64] & -- \\
        —    —    — Conv2d: 3-20 & [1, 128, 64, 64] & [3, 3] & [1, 128, 64, 64] & 147,456 \\
        —    —    — BatchNorm2d: 3-21 & [1, 128, 64, 64] & -- & [1, 128, 64, 64] & 256 \\
        —    —    — ReLU: 3-22 & [1, 128, 64, 64] & -- & [1, 128, 64, 64] & -- \\
        —    —    — Conv2d: 3-23 & [1, 128, 64, 64] & [3, 3] & [1, 128, 64, 64] & 147,456 \\
        —    —    — BatchNorm2d: 3-24 & [1, 128, 64, 64] & -- & [1, 128, 64, 64] & 256 \\
        —    —    — ReLU: 3-25 & [1, 128, 64, 64] & -- & [1, 128, 64, 64] & -- \\
        — Sequential: 1-7 & [1, 128, 64, 64] & -- & [1, 256, 32, 32] & -- \\
        —    — BasicBlock: 2-5 & [1, 128, 64, 64] & -- & [1, 256, 32, 32] & -- \\
        —    —    — Conv2d: 3-26 & [1, 128, 64, 64] & [3, 3] & [1, 256, 32, 32] & 294,912 \\
        —    —    — BatchNorm2d: 3-27 & [1, 256, 32, 32] & -- & [1, 256, 32, 32] & 512 \\
        —    —    — ReLU: 3-28 & [1, 256, 32, 32] & -- & [1, 256, 32, 32] & -- \\
        —    —    — Conv2d: 3-29 & [1, 256, 32, 32] & [3, 3] & [1, 256, 32, 32] & 589,824 \\
        —    —    — BatchNorm2d: 3-30 & [1, 256, 32, 32] & -- & [1, 256, 32, 32] & 512 \\
        —    —    — Sequential: 3-31 & [1, 128, 64, 64] & -- & [1, 256, 32, 32] & 33,280 \\
        —    —    — ReLU: 3-32 & [1, 256, 32, 32] & -- & [1, 256, 32, 32] & -- \\
        —    — BasicBlock: 2-6 & [1, 256, 32, 32] & -- & [1, 256, 32, 32] & -- \\
        —    —    — Conv2d: 3-33 & [1, 256, 32, 32] & [3, 3] & [1, 256, 32, 32] & 589,824 \\
        —    —    — BatchNorm2d: 3-34 & [1, 256, 32, 32] & -- & [1, 256, 32, 32] & 512 \\
        —    —    — ReLU: 3-35 & [1, 256, 32, 32] & -- & [1, 256, 32, 32] & -- \\
        —    —    — Conv2d: 3-36 & [1, 256, 32, 32] & [3, 3] & [1, 256, 32, 32] & 589,824 \\
        —    —    — BatchNorm2d: 3-37 & [1, 256, 32, 32] & -- & [1, 256, 32, 32] & 512 \\
        —    —    — ReLU: 3-38 & [1, 256, 32, 32] & -- & [1, 256, 32, 32] & -- \\
        — Sequential: 1-8 & [1, 256, 32, 32] & -- & [1, 512, 16, 16] & -- \\
        —    — BasicBlock: 2-7 & [1, 256, 32, 32] & -- & [1, 512, 16, 16] & -- \\
        —    —    — Conv2d: 3-39 & [1, 256, 32, 32] & [3, 3] & [1, 512, 16, 16] & 1,179,648 \\
        —    —    — BatchNorm2d: 3-40 & [1, 512, 16, 16] & -- & [1, 512, 16, 16] & 1,024 \\
        —    —    — ReLU: 3-41 & [1, 512, 16, 16] & -- & [1, 512, 16, 16] & -- \\
        —    —    — Conv2d: 3-42 & [1, 512, 16, 16] & [3, 3] & [1, 512, 16, 16] & 2,359,296 \\
        —    —    — BatchNorm2d: 3-43 & [1, 512, 16, 16] & -- & [1, 512, 16, 16] & 1,024 \\
        —    —    — Sequential: 3-44 & [1, 256, 32, 32] & -- & [1, 512, 16, 16] & 132,096 \\
        —    —    — ReLU: 3-45 & [1, 512, 16, 16] & -- & [1, 512, 16, 16] & -- \\
        —    — BasicBlock: 2-8 & [1, 512, 16, 16] & -- & [1, 512, 16, 16] & -- \\
        —    —    — Conv2d: 3-46 & [1, 512, 16, 16] & [3, 3] & [1, 512, 16, 16] & 2,359,296 \\
        —    —    — BatchNorm2d: 3-47 & [1, 512, 16, 16] & -- & [1, 512, 16, 16] & 1,024 \\
        —    —    — ReLU: 3-48 & [1, 512, 16, 16] & -- & [1, 512, 16, 16] & -- \\
        —    —    — Conv2d: 3-49 & [1, 512, 16, 16] & [3, 3] & [1, 512, 16, 16] & 2,359,296 \\
        —    —    — BatchNorm2d: 3-50 & [1, 512, 16, 16] & -- & [1, 512, 16, 16] & 1,024 \\
        —    —    — ReLU: 3-51 & [1, 512, 16, 16] & -- & [1, 512, 16, 16] & -- \\
        \bottomrule
      \end{tabular}
  }
  
  \label{tbl:arch_encoder}
\end{table*}

\begin{table*}
  \centering
  \caption{Decoder architecture.}
  
  \begin{tabular}{@{}lcccc@{}}
    \toprule
    Layer (type:depth-idx) & Input Shape & Kernel Shape & Output Shape & Param \# \\
    \bottomrule
    Decoder & [1, 64, 256, 256] & -- & [1, 12, 512, 512] & -- \\
    — Upsample: 1-1 & [1, 512, 16, 16] & -- & [1, 512, 32, 32] & -- \\
    — Sequential: 1-2 & [1, 768, 32, 32] & -- & [1, 256, 32, 32] & -- \\
    —    — Conv2d: 2-1 & [1, 768, 32, 32] & [3, 3] & [1, 256, 32, 32] & 1,769,728 \\
    —    — ReLU: 2-2 & [1, 256, 32, 32] & -- & [1, 256, 32, 32] & -- \\
    —    — Conv2d: 2-3 & [1, 256, 32, 32] & [3, 3] & [1, 256, 32, 32] & 590,080 \\
    —    — ReLU: 2-4 & [1, 256, 32, 32] & -- & [1, 256, 32, 32] & -- \\
    — Upsample: 1-3 & [1, 256, 32, 32] & -- & [1, 256, 64, 64] & -- \\
    — Sequential: 1-4 & [1, 384, 64, 64] & -- & [1, 128, 64, 64] & -- \\
    —    — Conv2d: 2-5 & [1, 384, 64, 64] & [3, 3] & [1, 128, 64, 64] & 442,496 \\
    —    — ReLU: 2-6 & [1, 128, 64, 64] & -- & [1, 128, 64, 64] & -- \\
    —    — Conv2d: 2-7 & [1, 128, 64, 64] & [3, 3] & [1, 128, 64, 64] & 147,584 \\
    —    — ReLU: 2-8 & [1, 128, 64, 64] & -- & [1, 128, 64, 64] & -- \\
    — Upsample: 1-5 & [1, 128, 64, 64] & -- & [1, 128, 128, 128] & -- \\
    — Sequential: 1-6 & [1, 192, 128, 128] & -- & [1, 64, 128, 128] & -- \\
    —    — Conv2d: 2-9 & [1, 192, 128, 128] & [3, 3] & [1, 64, 128, 128] & 110,656 \\
    —    — ReLU: 2-10 & [1, 64, 128, 128] & -- & [1, 64, 128, 128] & -- \\
    —    — Conv2d: 2-11 & [1, 64, 128, 128] & [3, 3] & [1, 64, 128, 128] & 36,928 \\
    —    — ReLU: 2-12 & [1, 64, 128, 128] & -- & [1, 64, 128, 128] & -- \\
    — Upsample: 1-7 & [1, 64, 128, 128] & -- & [1, 64, 256, 256] & -- \\
    — Sequential: 1-8 & [1, 128, 256, 256] & -- & [1, 64, 256, 256] & -- \\
    —    — Conv2d: 2-13 & [1, 128, 256, 256] & [3, 3] & [1, 64, 256, 256] & 73,792 \\
    —    — ReLU: 2-14 & [1, 64, 256, 256] & -- & [1, 64, 256, 256] & -- \\
    —    — Conv2d: 2-15 & [1, 64, 256, 256] & [3, 3] & [1, 64, 256, 256] & 36,928 \\
    —    — ReLU: 2-16 & [1, 64, 256, 256] & -- & [1, 64, 256, 256] & -- \\
    — Upsample: 1-9 & [1, 64, 256, 256] & -- & [1, 64, 512, 512] & -- \\
    — Conv2d: 1-10 & [1, 64, 512, 512] & [3, 3] & [1, 12, 512, 512] & 6,924 \\
    \bottomrule
  \end{tabular}
  
  \label{tbl:arch_decoder}
\end{table*}

\begin{table*}
  \centering
  \caption{Discriminator architecture.}
  
  \begin{tabular}{@{}lcccc@{}}
    \toprule
    Layer (type:depth-idx) & Input Shape & Kernel Shape & Output Shape & Param \# \\
    \bottomrule
    Discriminator & [1, 1, 256, 256] & -- & -- & -- \\
    — ModuleList: 1-1 & -- & -- & -- & -- \\
    —    — DiscriminatorBlock: 2-1 & [1, 1, 256, 256] & -- & [1, 32, 128, 128] & -- \\
    —    —    — Conv2d: 3-1 & [1, 1, 256, 256] & [1, 1] & [1, 32, 128, 128] & 64 \\
    —    —    — Sequential: 3-2 & [1, 1, 256, 256] & -- & [1, 32, 256, 256] & 9,568 \\
    —    —    — Sequential: 3-3 & [1, 32, 256, 256] & -- & [1, 32, 128, 128] & 9,248 \\
    —    — DiscriminatorBlock: 2-2 & [1, 32, 128, 128] & -- & [1, 64, 64, 64] & -- \\
    —    —    — Conv2d: 3-4 & [1, 32, 128, 128] & [1, 1] & [1, 64, 64, 64] & 2,112 \\
    —    —    — Sequential: 3-5 & [1, 32, 128, 128] & -- & [1, 64, 128, 128] & 55,424 \\
    —    —    — Sequential: 3-6 & [1, 64, 128, 128] & -- & [1, 64, 64, 64] & 36,928 \\
    —    — DiscriminatorBlock: 2-3 & [1, 64, 64, 64] & -- & [1, 128, 32, 32] & -- \\
    —    —    — Conv2d: 3-7 & [1, 64, 64, 64] & [1, 1] & [1, 128, 32, 32] & 8,320 \\
    —    —    — Sequential: 3-8 & [1, 64, 64, 64] & -- & [1, 128, 64, 64] & 221,440 \\
    —    —    — Sequential: 3-9 & [1, 128, 64, 64] & -- & [1, 128, 32, 32] & 147,584 \\
    —    — DiscriminatorBlock: 2-4 & [1, 128, 32, 32] & -- & [1, 128, 16, 16] & -- \\
    —    —    — Conv2d: 3-10 & [1, 128, 32, 32] & [1, 1] & [1, 128, 16, 16] & 16,512 \\
    —    —    — Sequential: 3-11 & [1, 128, 32, 32] & -- & [1, 128, 32, 32] & 295,168 \\
    —    —    — Sequential: 3-12 & [1, 128, 32, 32] & -- & [1, 128, 16, 16] & 147,584 \\
    —    — DiscriminatorBlock: 2-5 & [1, 128, 16, 16] & -- & [1, 128, 8, 8] & -- \\
    —    —    — Conv2d: 3-13 & [1, 128, 16, 16] & [1, 1] & [1, 128, 8, 8] & 16,512 \\
    —    —    — Sequential: 3-14 & [1, 128, 16, 16] & -- & [1, 128, 16, 16] & 295,168 \\
    —    —    — Sequential: 3-15 & [1, 128, 16, 16] & -- & [1, 128, 8, 8] & 147,584 \\
    —    — DiscriminatorBlock: 2-6 & [1, 128, 8, 8] & -- & [1, 128, 4, 4] & -- \\
    —    —    — Conv2d: 3-16 & [1, 128, 8, 8] & [1, 1] & [1, 128, 4, 4] & 16,512 \\
    —    —    — Sequential: 3-17 & [1, 128, 8, 8] & -- & [1, 128, 8, 8] & 295,168 \\
    —    —    — Sequential: 3-18 & [1, 128, 8, 8] & -- & [1, 128, 4, 4] & 147,584 \\
    —    — DiscriminatorBlock: 2-7 & [1, 128, 4, 4] & -- & [1, 128, 2, 2] & -- \\
    —    —    — Conv2d: 3-19 & [1, 128, 4, 4] & [1, 1] & [1, 128, 2, 2] & 16,512 \\
    —    —    — Sequential: 3-20 & [1, 128, 4, 4] & -- & [1, 128, 4, 4] & 295,168 \\
    —    —    — Sequential: 3-21 & [1, 128, 4, 4] & -- & [1, 128, 2, 2] & 147,584 \\
    —    — DiscriminatorBlock: 2-8 & [1, 128, 2, 2] & -- & [1, 128, 2, 2] & -- \\
    —    —    — Conv2d: 3-22 & [1, 128, 2, 2] & [1, 1] & [1, 128, 2, 2] & 16,512 \\
    —    —    — Sequential: 3-23 & [1, 128, 2, 2] & -- & [1, 128, 2, 2] & 295,168 \\
    — ModuleList: 1-2 & -- & -- & -- & -- \\
    — Conv2d: 1-3 & [1, 128, 2, 2] & [3, 3] & [1, 128, 2, 2] & 147,584 \\
    — Flatten: 1-4 & [1, 128, 2, 2] & -- & [1, 512] & -- \\
    — Linear: 1-5 & [1, 512] & -- & [1, 1] & 513 \\
    \bottomrule
  \end{tabular}
  
  \label{tbl:arch_discriminator}
\end{table*}

\end{document}